\documentclass[11pt,a4paper]{article}
\usepackage[numbers,sort&compress]{natbib}
\usepackage{amssymb}
\usepackage{mathtools}
\makeatletter
\def\input@path{{template/}}
\makeatother
\usepackage{tapri}
\usepackage{amsmath,amsfonts,bm}

\def\eqref#1{equation~\ref{#1}}
\def\1{\bm{1}}

\ifdefined\mathsfit\else
\DeclareMathAlphabet{\mathsfit}{\encodingdefault}{\sfdefault}{m}{sl}
\SetMathAlphabet{\mathsfit}{bold}{\encodingdefault}{\sfdefault}{bx}{n}
\fi

\usepackage{url}            % simple URL typesetting
\usepackage{booktabs}       % professional-quality tables
\usepackage{nicefrac}       % compact symbols for 1/2, etc.
\usepackage{lipsum}
\usepackage{adjustbox}
\usepackage{subcaption}
\usepackage{needspace}
\usepackage{afterpage}
\usepackage{xspace}
\usepackage{array}
\usepackage{arydshln}
\usepackage[detect-weight=true]{siunitx}
\usepackage{pgfplotstable}
\pgfplotsset{compat=1.18}
\usepackage{cleveref}
\crefname{appendix}{appendix}{appendices}
\Crefname{appendix}{Appendix}{Appendices}

\definecolor{tablecheck}{HTML}{5FAF82}
\definecolor{tablecross}{HTML}{D96F63}
\newcommand{\cmark}{\textcolor{tablecheck}{\ensuremath{\checkmark}}}
\newcommand{\xmark}{\textcolor{tablecross}{\ensuremath{\times}}}

\newcount\bestvaluerow
\newcommand{\autoboldbest}[3]{%
  \ifnum\pgfplotstablerow>-1\relax
  \pgfplotstablegetelem{0}{\pgfplotstablecolname}\of{#2}%
  \edef\bestcolumnvalue{\pgfplotsretval}%
  \bestvaluerow=1\relax
  \loop\ifnum\bestvaluerow<\pgfplotstablerows\relax
  \pgfplotstablegetelem{\the\bestvaluerow}{\pgfplotstablecolname}\of{#2}%
  \def\bestdirection{#1}%
  \def\minimumdirection{min}%
  \ifx\bestdirection\minimumdirection
  \pgfmathparse{min(\bestcolumnvalue,\pgfplotsretval)}%
  \else
  \pgfmathparse{max(\bestcolumnvalue,\pgfplotsretval)}%
  \fi
  \edef\bestcolumnvalue{\pgfmathresult}%
  \advance\bestvaluerow by 1\relax
  \repeat
  \let\secondcolumnvalue\empty
  \bestvaluerow=0\relax
  \loop\ifnum\bestvaluerow<\pgfplotstablerows\relax
  \pgfplotstablegetelem{\the\bestvaluerow}{\pgfplotstablecolname}\of{#2}%
  \pgfmathparse{abs((\pgfplotsretval)-(\bestcolumnvalue)) >= 0.0001}%
  \ifdim\pgfmathresult pt>0pt
  \ifx\secondcolumnvalue\empty
  \edef\secondcolumnvalue{\pgfplotsretval}%
  \else
  \def\bestdirection{#1}%
  \def\minimumdirection{min}%
  \ifx\bestdirection\minimumdirection
  \pgfmathparse{min(\secondcolumnvalue,\pgfplotsretval)}%
  \else
  \pgfmathparse{max(\secondcolumnvalue,\pgfplotsretval)}%
  \fi
  \edef\secondcolumnvalue{\pgfmathresult}%
  \fi
  \fi
  \advance\bestvaluerow by 1\relax
  \repeat
  \pgfmathparse{abs((#3)-(\bestcolumnvalue)) < 0.0001}%
  \ifdim\pgfmathresult pt>0pt
  \pgfkeysalso{/pgfplots/table/@cell content/.add={\bfseries }{}}%
  \else
  \ifx\secondcolumnvalue\empty
  \else
  \pgfmathparse{abs((#3)-(\secondcolumnvalue)) < 0.0001}%
  \ifdim\pgfmathresult pt>0pt
  \pgfkeyssetvalue{/pgfplots/table/@cell content}{{\underline{#3}}}%
  \fi
  \fi
  \fi
  \fi
}
\pgfplotstableset{
  best/.style 2 args={
    postproc cell content/.append code={\autoboldbest{#1}{#2}{##1}}
  }
}

\newcommand{\methodname}[0]{\mbox{BRAID}\xspace}
\newcommand{\methodnamefull}[0]{Bilevel Representations for Agent Interaction Dynamics\xspace}

\usepackage{tikz}
\usetikzlibrary{arrows.meta, calc}

\taprisetup{layout=wide,teaser-max-height=55mm,
  running-title={Generative Interactions · Tapri Lab · Preprint}}

\title{Generative Interactions: Weaving Multiparty Human Motion with Bilevel Latent Dynamics}
\tapriauthor{Ojas Shirekar}{1}
\tapriauthor{Yash Surange}{1}
\tapriauthor{Augustinas Jučas}{2}
\tapriauthor{Chirag Raman}{1}
\tapriaffiliation{1}{Delft University of Technology, Delft, The Netherlands}
\tapriaffiliation{2}{University of Oxford, Oxford, United Kingdom}
\tapricontact{\href{mailto:o.k.shirekar@tudelft.nl}{o.k.shirekar@tudelft.nl},
  \href{mailto:y.surange@student.tudelft.nl}{y.surange@student.tudelft.nl},
  \href{mailto:augustinasjucas@gmail.com}{augustinasjucas@gmail.com},
  \href{mailto:c.a.raman@tudelft.nl}{c.a.raman@tudelft.nl}}

\tapriabstract{Human social behaviour is not a collection of independent motions, but a jointly organised
  process in which group dynamics and individual variation continuously shape one another. Yet existing
  social motion models often prioritise plausible trajectories while leaving interaction state implicit,
  limiting their ability to transfer across groups, tasks, and partial-observation regimes. To address this
  gap, we introduce \methodnamefull (\methodname), a hierarchical sequential latent-variable model for
  generative multi-person interaction. \methodname explicitly formulates social motion generation as a
  meta-transfer learning problem: shared interaction priors are learned across datasets and adapted through
  arbitrary context sets of observed people and joints. The model represents each scene through a
  group-level latent state that captures shared interaction dynamics and person-level latent states that
  capture individual behaviour conditioned on the evolving group context. This modelling choice enables
  coherent generation under full, sparse, or partial observations while exposing compact social-state
  vectors that can serve as an interface for downstream embodied-agent systems. We evaluate \methodname
  under a unified SMPL-based representation on social forecasting, tracking and in-filling, and response
  generation, using metrics that assess not only reconstruction accuracy but also realism, diversity,
  temporal alignment, and interpersonal coordination. We further analyse the hierarchical latent space,
  showing that it captures separable group- and individual-level structure.}
\tapriteaser[fig:teaser]{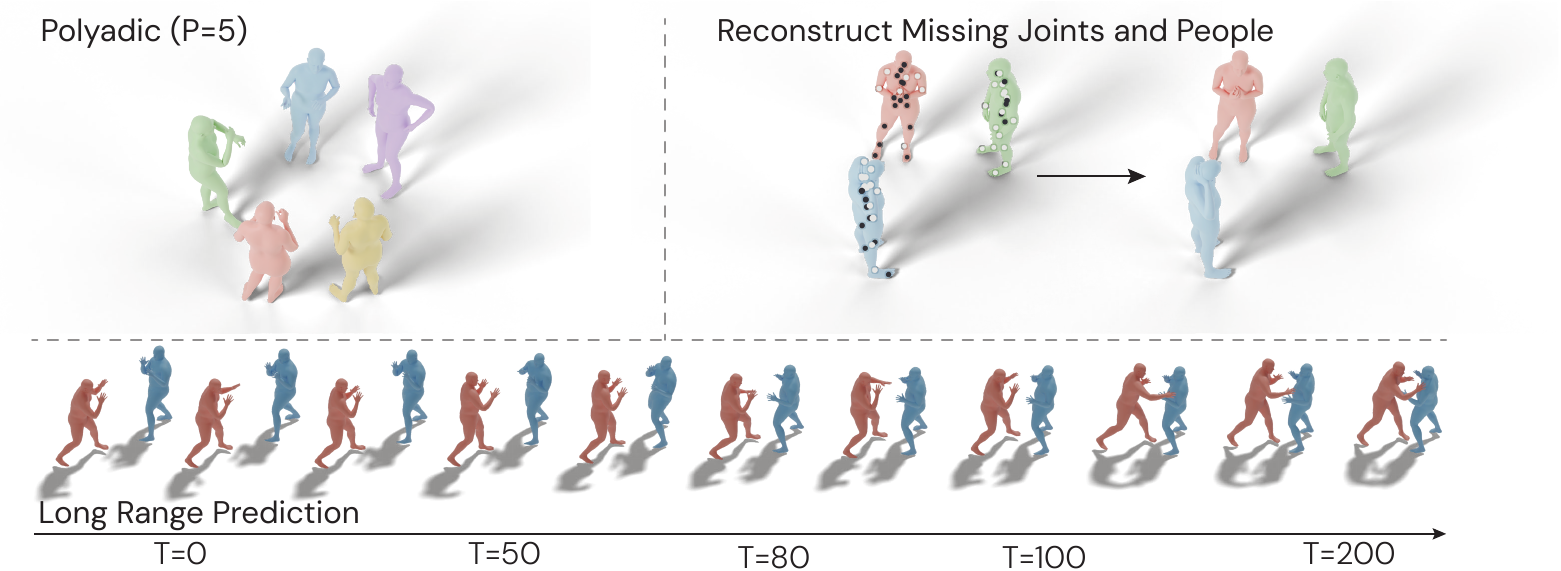}{\textbf{A generative model for multi-agent interaction.}
  We propose \methodnamefull (\methodname), a unified model of multi-party group behaviour.}

\begin{document}

\maketitle

\section{Introduction}\label{sec:intro}

Human social interaction couples group-level coordination with individual behavioural variation. Embodied AI
systems that perceive or participate in these interactions must track how a situation is unfolding, how
participants influence one another, and which continuations are plausible. This motivates learning structured
social-state representations alongside realistic motion generation. These representations must account for
non-verbal signals, including posture, head nods, gestures, and
gaze~\citep{Raman2023xgd, Kleef2007, Pruitt2011, Smith2009, Zanlungo2017, Matsumoto2007,
Terry1999, Rathbone2023, Oers2005, Levinson2015, Sacks1974, Markhorst2026}, while accommodating multiple valid
responses shaped by individual traits, group roles, cultural contexts, and situational norms.

Learning transferable social structure is difficult with scarce and fragmented interaction data. Motion
capture datasets such as Panoptic~\citep{Joo2016}, DnD~\citep{Mughal2024}, and
Embody3D~\citep{McLean2025} cover only a fraction of real-world situations and interaction styles.
As~\citet{Raman2023xgd} argue, the challenge is therefore not only data scarcity but transfer: models must
learn structure that generalises across datasets, contexts, and observation regimes.

Existing human motion generators~\citep{Tevet2022dzu, Tevet2022h1j, Guo2022, Diomataris2024} provide strong
foundations for individual movement. Social motion methods address interaction, but often leave the interplay
between shared dynamics and individual traits implicit~\citep{Wang2021xhw,Chew2025,Lin2026}, with limited
adaptability to varying context. The central gap is a reusable social-state representation that jointly
captures group and individual dynamics, supports generation under full or partial observations, and transfers
across datasets. These capabilities are rarely addressed together: plausible trajectories remain the primary
endpoint, while compact, inspectable state representations could also inform policies, planners, and memory
systems that condition on ongoing interaction.

To address this gap, we propose \methodname (\textit{\methodnamefull}), a hierarchical sequential
latent-variable model for multi-person social behaviour. A group latent captures shared interaction dynamics,
while person latents encode individual behaviour conditioned on the group state. Their coupled temporal
evolution supports coherent generation, and arbitrary context sets accommodate full, sparse, or partial
observations. We learn shared priors through cross-dataset meta-transfer learning across conversational,
dance, and boxing data. The resulting group and person latents expose compact, inspectable social-state vectors
for downstream agents.

% The teaser figure (fig:teaser) lives in the opening panel; see \tapriteaser in main.tex.

Our main contributions can be summarised as:
(i) A generative state space model for multi-person social behaviour under full, partial, or absent
observations. (ii) A hierarchical social-state representation of shared group dynamics and individual
behavioural variation.
(iii) Evaluation across datasets and context regimes, including partially observed interaction partners in a
scene, and (iv) analysis of group- and person-level latent structure, informed by cognitive science.

\section{Related Work}\label{sec:related-work}

\begin{table}[htbp]
  \centering
  \caption{Comparison of methods across interaction capabilities. Latent space denotes an explicit compact
  latent representation of behaviour; polyadic denotes support for \(P \geq 3\) participants.}
  \label{tab:comparison}
  \footnotesize
  \renewcommand{\arraystretch}{1.1}
  \setlength{\tabcolsep}{1.5pt}
  \newcommand{\rwhead}[2]{\adjustbox{max width=\rwcapabilitywidth}{\shortstack[c]{\textbf{#1}\\\textbf{#2}}}}
  % Calculate equal capability columns without tabularx, which conflicts with
  % arydshln in the June 2026 LaTeX release used by Overleaf.
  \newlength{\rwmethodwidth}
  \settowidth{\rwmethodwidth}{Social Processes~\citeyearpar{Raman2023xgd}}
  \edef\rwcapabilitywidth{\the\dimexpr(\textwidth-\rwmethodwidth-16\tabcolsep)/8\relax}
  \begin{tabular}{@{}l *{8}{>{\centering\arraybackslash}p{\rwcapabilitywidth}}@{}}
    \toprule
    \textbf{Method}
    & \rwhead{Missing/}{Noisy data}
    & \rwhead{Latent}{space}
    & \rwhead{Partner}{inpaint}
    & \rwhead{Partner}{predict}
    & \rwhead{Long}{motion}
    & \rwhead{Agentic}{generation}
    & \rwhead{Joint future}{prediction}
    & \rwhead{Polyadic}{\(P \geq 3\)} \\
    \midrule
    DuoLando~\citeyearpar{Siyao2024}   & \xmark & \xmark & \cmark & \xmark & \xmark & \xmark & \xmark & \xmark \\
    ReMoS~\citeyearpar{Ghosh2023}      & \xmark & \xmark & \cmark & \xmark & \xmark & \xmark & \xmark & \xmark \\
    ReGenNet~\citeyearpar{Xu2024-xs}     & \xmark & \xmark & \cmark & \xmark & \xmark & \xmark & \xmark & \xmark \\
    Human-X~\citeyearpar{Ji2025}  & \xmark & \xmark & \cmark & \xmark & \xmark & \xmark & \xmark & \xmark \\
    ARFlow~\citeyearpar{Jiang2025}  & \xmark & \xmark & \xmark & \cmark & \cmark & \xmark & \xmark & \xmark \\
    Ready-to-React~\citeyearpar{Cen2025}  & \xmark & \xmark & \xmark & \cmark & \cmark & \xmark & \xmark & \xmark \\
    MAGNet~\citeyearpar{Maluleke2025}     & \xmark & \xmark & \cmark & \cmark & \cmark & \cmark & \cmark & \cmark \\
    Social Processes~\citeyearpar{Raman2023xgd}  & \xmark & \cmark & \xmark & \cmark & \xmark & \xmark & \cmark
    & \cmark \\
    \cmidrule{1-9}
    \textbf{\methodname}                       & \cmark & \cmark & \cmark & \cmark & \cmark & \cmark & \cmark
    & \cmark \\
    \bottomrule
  \end{tabular}
\end{table}

\paragraph{Human motion generation.} Recent motion generation models produce high-quality single-person
motions from text, action labels, or goals~\citep{Tevet2022h1j,Tevet2022dzu,Guo2022,Diomataris2024}.
These systems provide strong priors for individual behaviour, but they generally treat the person as the
unit of generation. Social settings require a different abstraction: a pause, gesture, or step is meaningful
because it is coordinated with other participants. \methodname builds on this progress but models the
interaction, rather than an isolated body.

\paragraph{Social motion generation.} Social motion methods explicitly target interaction through
multi-person prediction, inpainting, response generation, or specialised domains such as gesture, dance, and
reactive character motion~\citep{Wang2021xhw,Tanke2023,Maluleke2025,Mughal2024,Siyao2024,Cen2025,Chew2025}.
Adjacent to this literature, Social Processes~\citep{Raman2023xgd} forecast nonverbal social cues in
conversational groups by meta-learning group-specific dynamics, but do not address full-body social motion
generation. Together, these approaches have substantially broadened the modelling of social interaction, but
most still treat plausible future cues or trajectories as the primary endpoint. Group-level state is either
absent, implicit in symmetric multi-agent architectures, or embedded inside a denoising process that does not
naturally expose a compact representation for inspection or downstream
integration~\citep{Preechakul2021,Pandey2022,Chen2022o2e,Tanke2025-wq}.
\methodname instead learns explicit group and person-level latents, \(z^g_t\) and \(z^p_t\), so that shared
interaction dynamics can condition individual behaviour while remaining available as reusable social-state
representations.

\paragraph{Neural Processes and meta-learning.} Neural Processes model distributions over functions using a
latent variable conditioned on a context set~\citep{Garnelo2018,Kim2019l993,Garnelo2018z2k}. This makes
them well suited to few-shot generalisation and arbitrary missing-observation patterns. Sequential Neural
Processes add temporal dependence between latent states~\citep{Singh2019-qw}, and Social
Processes~\citep{Raman2023xgd} show that this family can model conversational group dynamics from sparse
interaction data. \methodname extends this line by replacing the single latent state with a temporal
hierarchy: a group latent for shared interaction structure and person latents for individual behaviour
conditioned on that structure.

\paragraph{Hierarchical latent variable models.} Hierarchical VAEs use top-down structure to let higher-level
latents shape lower-level ones, often yielding richer multi-scale representations~\citep{Snderby2016}.
Sequential variants couple these hierarchies to recurrent state models~\citep{Gregor2018}, and
precision-weighted or product-of-experts posteriors combine bottom-up evidence with top-down
priors~\citep{Hinton2002,Snderby2016}. \methodname instantiates these ideas in the social motion domain:
the higher-level latent captures emergent interaction properties, while person-level latents encode
individual variation conditioned on that shared context. Our staged KL annealing and free-bits terms follow
established practice for stabilising hierarchical VAEs~\citep{Kingma2016,Vahdat2020,Snderby2016}, with a
warm-up order chosen to stabilise group structure before person-level variation.

\section{Problem Statement}\label{sec:problem-statement}

\begin{figure*}[t]
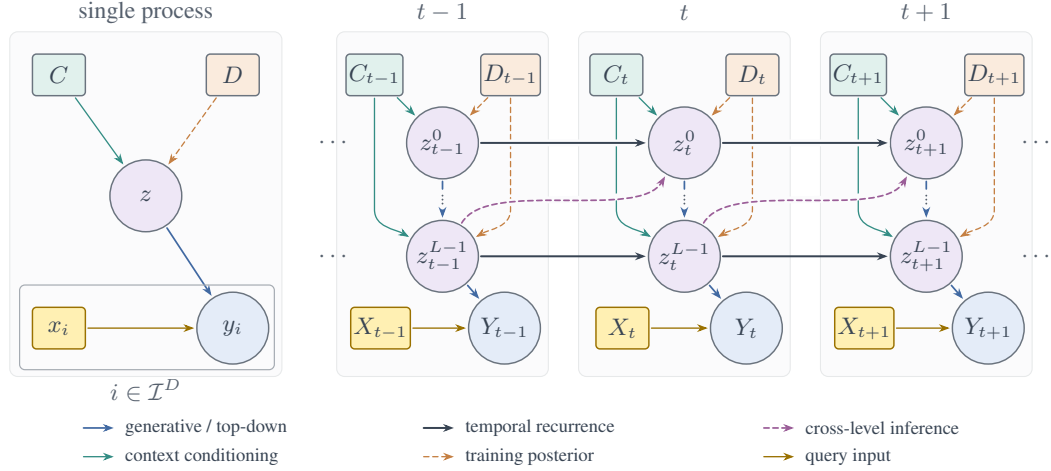

  \centering
  % Box the diagrams to prevent incidental input whitespace from wrapping a line.
  \newsavebox{\npcomparisonbox}
  \newsavebox{\hierarchycomparisonbox}
  \sbox{\npcomparisonbox}{\def\npScale{1}% Standard Neural Process. Styles and legend are shared with plate_diagram.
\providecommand{\npScale}{1}%
\input{figs/process_diagram_styles}%
\begin{tikzpicture}[scale=\npScale, transform shape]
  \path[use as bounding box] (-0.1,-1.0) rectangle (3.7,4.35);
  \draw[process panel] (0,-0.64) rectangle (3.6,3.92);
  \node[process label] at (1.8,4.18) {single process};
  \node[process context] (C) at (0.65,3.35) {\(C\)};
  \node[process target] (D) at (2.95,3.35) {\(D\)};
  \node[process latent] (z) at (1.8,1.75) {\(z\)};
  \node[process query] (X) at (0.65,0) {\(x_i\)};
  \node[process output] (Y) at (2.95,0) {\(y_i\)};
  % Automatic border intersections keep each arrow radial to its destination.
  \draw[process conditioning] (C) -- (z);
  \draw[process posterior] (D) -- (z);
  \draw[process generative] (z) -- (Y);
  \draw[process query edge] (X.east) -- (Y.west);
  \draw[draw=processInk!45, rounded corners=2pt, line width=0.45pt]
    (0.13,-0.55) rectangle (3.50,0.57);
  \node[process label] at (1.8,-0.82) {\(i\in\mathcal{I}^{D}\)};
\end{tikzpicture}%
}%
  \sbox{\hierarchycomparisonbox}{\def\plateScale{1}% Hierarchical sequential Neural Process, with one column per time step.
\providecommand{\plateScale}{1}%
\input{figs/process_diagram_styles}%
\begin{tikzpicture}[scale=\plateScale, transform shape]
  \path[use as bounding box] (-0.15,-1.0) rectangle (9.55,4.35);
  \foreach \i/\tsub in {0/{t-1},1/{t},2/{t+1}} {
    \pgfmathsetmacro{\px}{1.5+3.2*\i}
    \draw[process panel] (\px-1.40,-0.64) rectangle (\px+1.40,3.92);
    \node[process label] at (\px,4.18) {\(\tsub\)};
    \node[process context] (C\i) at (\px-0.90,3.35) {\(C_{\tsub}\)};
    \node[process target] (D\i) at (\px+0.90,3.35) {\(D_{\tsub}\)};
    \node[process latent] (z0_\i) at (\px,2.45) {\(z^{0}_{\tsub}\)};
    \node[process latent] (zL_\i) at (\px,0.95) {\(z^{L-1}_{\tsub}\)};
    \node[process query] (X\i) at (\px-0.82,0) {\(X_{\tsub}\)};
    \node[process output] (Y\i) at (\px+0.82,0) {\(Y_{\tsub}\)};
  }
  % Symmetric side lanes keep conditioning separate from the latent hierarchy.
  \foreach \i in {0,1,2} {
    \draw[process conditioning] (C\i) -- (z0_\i);
    \draw[process conditioning] (C\i.south)
      -- ($(C\i.south |- zL_\i.center)+(0,0.62)$)
      to[out=-90,in=150] (zL_\i.150);
    \draw[process posterior] (D\i) -- (z0_\i);
    \draw[process posterior] (D\i.south)
      -- ($(D\i.south |- zL_\i.center)+(0,0.62)$)
      to[out=-90,in=30] (zL_\i.30);
  }
  % Continuous temporal links cross side lanes with a small background clearance.
  \foreach \a/\b in {0/1,1/2} {
    \draw[draw=processPanelFill,line width=2.6pt] (z0_\a) -- (z0_\b);
    \draw[process temporal] (z0_\a) -- (z0_\b);
    \draw[process temporal] (zL_\a) -- (zL_\b);
    \draw[draw=processPanelFill,line width=2.6pt]
      (zL_\a.60) to[out=60,in=240,looseness=0.65] (z0_\b.240);
    \draw[process inference]
      (zL_\a.60) to[out=60,in=240,looseness=0.65] (z0_\b.240);
  }
  \foreach \i in {0,1,2} {
    % Split the vertical shaft around centered dots instead of drawing through them.
    \coordinate (levelgap\i) at ($(z0_\i.south)!0.5!(zL_\i.north)$);
    \draw[process generative,-] (z0_\i.south) -- ($(levelgap\i)+(0,3.3pt)$);
    \draw[process generative] ($(levelgap\i)+(0,-3.3pt)$) -- (zL_\i.north);
    \foreach \dotshift in {-1.5pt,0pt,1.5pt} {
      \fill[processInk!85] ($(levelgap\i)+(0pt,\dotshift)$) circle[radius=0.35pt];
    }
    \draw[process generative] (zL_\i) -- (Y\i);
    \draw[process query edge] (X\i.east) -- (Y\i.west);
  }
  \foreach \yy in {0.95,2.45} {
    \node[process label] at (0.05,\yy) {\(\cdots\)};
    \node[process label] at (9.35,\yy) {\(\cdots\)};
  }
\end{tikzpicture}%
}%
  \begin{minipage}[t]{0.28\textwidth}
    \vspace{0pt}
    \centering
    \noindent\usebox{\npcomparisonbox}\par
  \end{minipage}\hfill
  \begin{minipage}[t]{0.70\textwidth}
    \vspace{0pt}
    \centering
    \noindent\usebox{\hierarchycomparisonbox}\par
  \end{minipage}
  \par\smallskip
  % One compact legend for both diagrams.
\input{figs/process_diagram_styles}%
\begin{tikzpicture}[x=1cm,y=1cm]
  \foreach \xx/\yy/\style/\label in {
    0/0/process generative/{generative / top-down},
    4.5/0/process temporal/{temporal recurrence},
    9/0/process inference/{cross-level inference},
    0/-0.38/process conditioning/{context conditioning},
    4.5/-0.38/process posterior/{training posterior},
    9/-0.38/process query edge/{query input}} {
    \draw[\style] (\xx,\yy) -- (\xx+0.45,\yy);
    \node[anchor=west,font=\scriptsize,text=processInk,inner sep=2pt]
      at (\xx+0.5,\yy) {\label};
  }
\end{tikzpicture}
  \par\vspace{4pt}\nointerlineskip
  % Keep the shared legend with the artwork, before either subcaption.
  \begin{subfigure}[t]{0.28\textwidth}
    \captionsetup{skip=0pt}
    \caption{Neural Process: one latent for the observed context.}
    \label{fig:np-model}
  \end{subfigure}\hfill
  \begin{subfigure}[t]{0.70\textwidth}
    \captionsetup{skip=0pt}
    \caption{\methodname: a hierarchy of latents linked across time.
    Each column is one time step; vertical dots indicate intermediate levels.}
    \label{fig:shinp-model}
  \end{subfigure}
  \caption{Neural Process and \methodname. \(C\): observed context; \(D\supseteq C\): full target set;
  \(X\): query inputs; \(Y\): outputs. Dashed arrows denote inference-only dependencies.}
  \label{fig:gen-model}
\end{figure*}

Consider a temporal Neural Process setting with stochastic processes
\(\mathcal{P}_1, \dots, \mathcal{P}_T\). At each time step \(t\), we observe a possibly
empty context set \(C_t \coloneqq \{(x_t^i, y_t^i)\}_{i \in \mathcal{I}^C_t}\) and
define a target set \(D_t \coloneqq \{(x_t^i, y_t^i)\}_{i \in \mathcal{I}^D_t}\), with
\(\mathcal{I}^C_t \subseteq \mathcal{I}^D_t\). Let
\(X_t \coloneqq \{x_t^i\}_{i \in \mathcal{I}^D_t}\) and
\(Y_t \coloneqq \{y_t^i\}_{i \in \mathcal{I}^D_t}\) denote the corresponding target
inputs and outputs; the held-out elements \(D_t \setminus C_t\) are the query locations
and target values to be predicted.

A standard Neural Process uses a single latent \(z_t\) conditioned on the current
context \(C_t\), but does not model dependencies between successive processes. \methodname
instead introduces \(L\) latent levels \(z_t^{(0)}, \dots, z_t^{(L-1)}\), conditioned on
the context, latent history, and lower levels in the hierarchy. Writing \(C,D,X,Y\) for
the full temporal sequences, \(Z^{(k)} \coloneqq (z^{(k)}_1, \dots, z^{(k)}_T)\), and
\(Z_{<t} \coloneqq \{z^{(k)}_{<t}\}_{k=0}^{L-1}\), the generative model factorises as:
{\small
  \begin{align}\label{eqn:gen-model}
    &p(Y, Z^{(0)}, Z^{(1)}, \dots, Z^{(L-1)} \mid X, C) = \notag \\
    &\quad \prod_{t=1}^T p_\theta(Y_t \mid X_t, z_t^{(0)}, \dots, z_t^{(L-1)}) \,
    p_\theta(z_t^{(0)} \mid Z_{<t}, C_t)
    \prod_{k=1}^{L-1} p_\theta(z_t^{(k)} \mid Z_{<t}, z_t^{(k-1)}, C_t).
\end{align}}

The decoder likelihood factorises over target elements, and we initialise
\(z^{(k)}_0 \coloneqq \operatorname{null}\) for all \(k \in [0, L-1]\). In the two-level
social-motion model introduced in \Cref{sec:method}, \(z_t^{(0)}\) is the group latent
and \(z_t^{(1)}\) is the structured collection of person-level latents.

\section{Method}\label{sec:method}

% Defer until the current page is complete, then allow top placement only.
\afterpage{%
  \begin{figure}[!t]
    \centering
    \includegraphics[width=\linewidth]{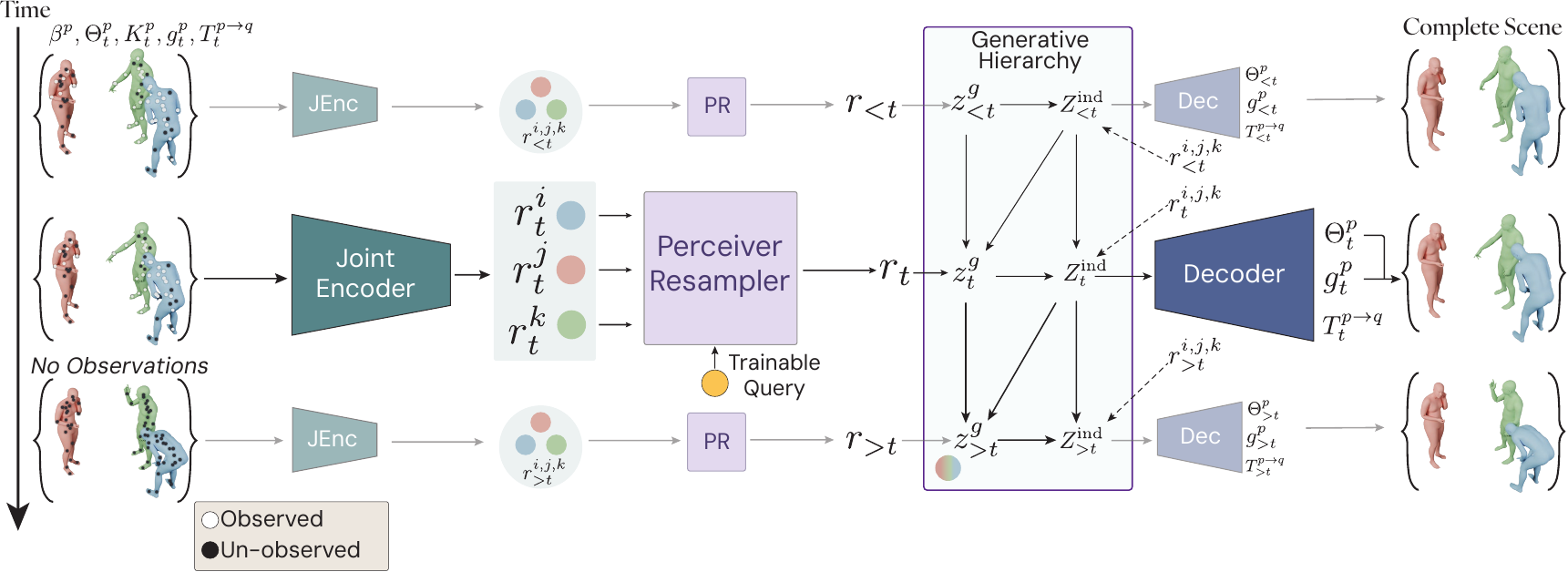}
    \caption{High-level architectural view of the model. Illustration shows a interacting triad. On the left
      the model receives partial observations at various time steps and it generates the complete scene
    including each person \(p \in \{i,j,k\}\)}
    \label{fig:arch-diagram}
  \end{figure}
}

While \methodname admits a \(L\)-level hierarchy, our approach instantiates the simplest
case of a two-level latent hierarchy. A group latent \(z_t^g\) captures shared
interaction state, such as phase, tempo, and proxemics, while person latents \(\{z_t^p\}_{p=1}^{P}\)
capture individual behaviour conditioned on that group context. This mirrors accounts of social interaction
as a jointly maintained process in which group-level coordination constrains individual
action~\citep{Sacks1974,Ptschulat2026,Hoehl2020,Chartrand1999,Hale2020}. We describe the scene representation,
latent dynamics, context encoder, and learning objective below.

\subsection{Motion Representation}\label{ssec:motion-repr}

We represent a scene as \(T\) time steps with \(P\) people. Each person has SMPL-X shape
\(\beta^p\in\mathbb{R}^{16}\), 6D joint rotations
\(\Theta_t^p\in\mathbb{R}^{J\times 6}\)~\citep{Zhou2018}, and root-relative joints
\(\tilde K_{t,j}^p\). We additionally store canonical root motion \(g_t^p\), using floor-anchored,
yaw-normalised SE(3) quantities similar to~\citet{Yi2024,Maluleke2025, Holden2016,Holden2017}. To retain
spatial awareness, we
encode pairwise canonical-frame transforms \(\mathbf{T}_t^{p\to q}\in\mathbb{R}^9\) with an MLP and mean-pool
them over observed partners before fusing them into the context representation.
Throughout, we write \(\mathcal{M}=\{\Delta\mathbf{T}^{\mathrm{can}},\,
\mathbf{T}^{\mathrm{can}\to\mathrm{root}},\, \mathbf{T}^{p\to q}\}\) for the three canonical transforms the
model predicts---the frame-to-frame canonical displacement, the canonical-to-root transform, and the
pairwise partner transform---each carrying a rotation \(R_m\) and a translation \(t_m\).

Following \cref{sec:problem-statement}, each observation is represented as a context point
\((x_t^{(p,j)}, y_t^{(p,j)})\), where \(x_t^{(p,j)}\) identifies a person and joint/root slot. For
\(j\geq1\), \(y_t^{(p,j)}=[\Theta^p_{t,j},\tilde K^p_{t,j}]\); for \(j=0\), \(y_t^{(p,0)}=g_t^p\). We define
the complete target set and observed context set using index sets
\(\mathcal{I}^D_t=[1,P]\times[0,J]\) and \(\mathcal{I}^C_t\subseteq\mathcal{I}^D_t\):
\(
  D_t \coloneqq \{(x_t^{(p,j)}, y_t^{(p,j)})\}_{(p,j)\in\mathcal{I}^D_t},
  C_t \coloneqq \{(x_t^{(p,j)}, y_t^{(p,j)})\}_{(p,j)\in\mathcal{I}^C_t}.
\)
Missing joints or people are simply omitted from \(C_t\), so forecasting, tracking/in-filling, and response
generation differ only in which indices are observed.

\subsection{Latent Encoding and Generative Model}\label{ssec:gen-model}

We instantiate the hierarchy from \cref{eqn:gen-model} with a group latent \(z_t^g\) and person latents
\(Z_t^{\mathrm{ind}}=\{z_t^p\}_{p=1}^P\). Given \(C_t\), the model samples the group latent, samples
person latents conditioned on it, and decodes all target motion quantities (dropping index \(j\) for simplicity):
\begin{align}
  p_\theta(Y, Z^g, Z^{\mathrm{ind}} \mid X, C)
  &= \prod_{t=1}^T
  p_\theta(z_t^g \mid z_{<t}^g, Z_{<t}^{\mathrm{ind}}, C_t) \nonumber \\
  &\quad \prod_{p=1}^P p_\theta(z_t^p \mid z_{<t}^p, z_t^g, C_t)\,
  p_\theta(y_t^p \mid x_t^p, z_t^{p}),
  \label{eqn:soc-gen-model}
\end{align}
The decoder factorises over target indices \((p,j)\in\mathcal{I}^D_t\), predicting \(g_t^p\) for \(j=0\)
and root-relative rotations/keypoints for \(j\geq1\).

\paragraph{Pretrained motion decoder.}
Body pose is decoded through a temporal motion VAE~\citep{Kingma2013} that is pretrained on complete motion
and then frozen. A trainable adapter maps the latents, person histories, and canonical root motion to the
VAE's codes (one per four frames), which the frozen decoder turns into joint rotations; separate heads
predict the canonical and partner transforms. The VAE thus supplies short-horizon body kinematics, while
\methodname{} models the social structure that selects among them. Each decoded frame therefore depends on a
short window of \(z^p\) and \(z^g\) around \(t\) (\Cref{appx:model-implementation}).

\subsection{Context Encoding}\label{ssec:context-enc}

Both the generative and inference networks condition on deterministic context representations: a group
summary \(r_t\) and per-person summaries \(r_t^p\). We embed observed joint features with joint tokens, use
cross-attention to relate context joints to target slots within each person, and add a continuous
RoPE-inspired person embedding~\citep{Su2021} so person slots remain distinguishable under masking.
Temporal attention and cross-person attention then capture short-range dynamics and coordination, and a
perceiver-style resampler~\citep{Jaegle2021} compresses per-person tokens into the group representation
\(r_t\).

\subsection{Inference and Learning}\label{ssec:inference}

Learning requires posterior inference over the latent social state implied by \cref{eqn:soc-gen-model}.
Since the true posterior \(p(Z^g, Z^{\mathrm{ind}} \mid C, D)\) is intractable, we train with a variational
approximation~\citep{Feynman1955,Peterson1987,Feynman1972,Blei2016,Kingma2013,Rezende2014}. We introduce an
amortised variational
distribution that mirrors the temporal and hierarchical structure of the generative model:
\begin{equation}
  q_\phi(Z^g, Z^{\mathrm{ind}} \mid C, D)
  = \prod_{t=1}^T q_\phi(z_t^g \mid z_{<t}^g, Z^{\mathrm{ind}}_{<t}, C, D)\,
  \prod_{p=1}^P q_\phi(z_t^p \mid z_{<t}^p, z_t^g, C, D).
  \label{eqn:soc-approx-posterior}
\end{equation}
% Figure 4/5 (experiments) defined here so that it lands at the top of page 6.
\begin{figure}[!t]
  \centering
  % Resolve the shared height before the minipages change the text width.
  \edef\motionfigureheight{\the\dimexpr0.3245\linewidth\relax}
  \begin{minipage}[t]{0.75\linewidth}
    \vspace{0pt}
    \centering
    \includegraphics[height=\motionfigureheight]{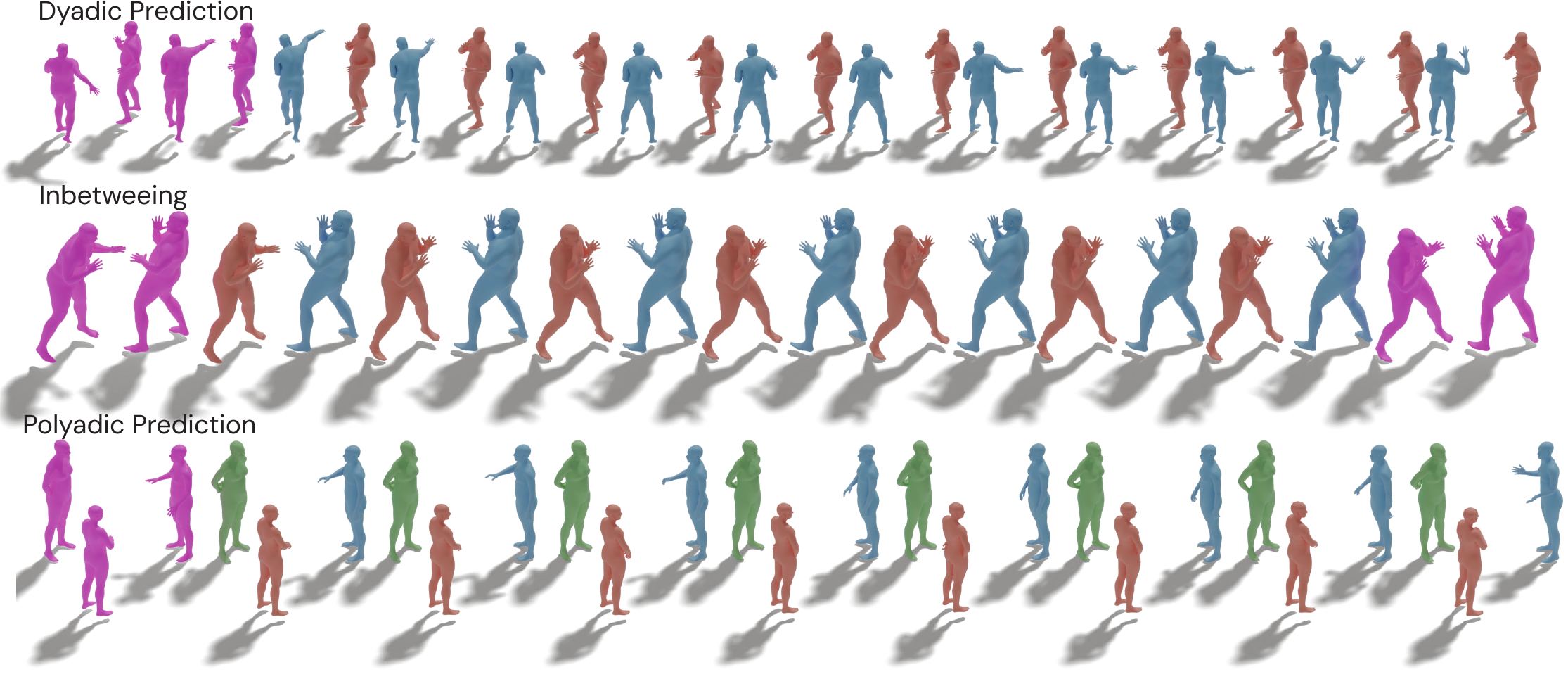}
    \caption{Samples from \methodname for dyadic prediction, in-betweening, and polyadic prediction,
    spanning boxing and conversational interactions. Frames in \textcolor[HTML]{B747A9}{Pink} are given as input. Qualitative videos are in the Supplement.}
    \label{fig:rollouts-diagram-1}
  \end{minipage}
  \hfill
  \begin{minipage}[t]{0.22\linewidth}
    \vspace{0pt}
    \centering
    \includegraphics[height=\motionfigureheight]{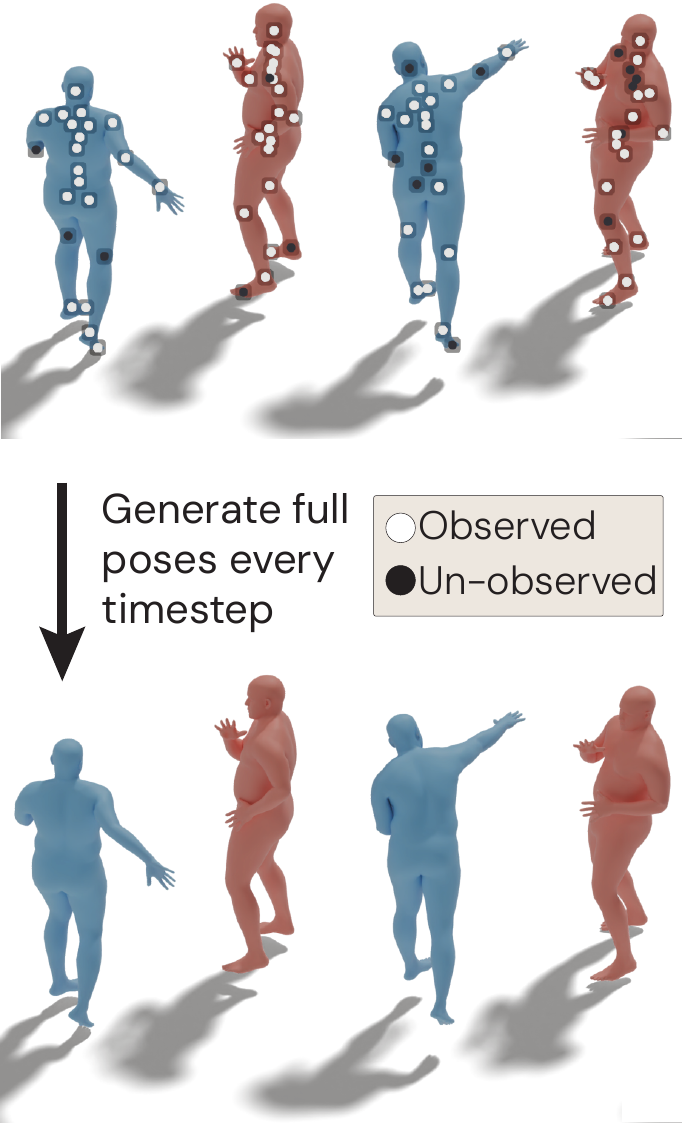}
    \captionsetup{justification=raggedright,singlelinecheck=false}
    \caption{Full-pose generation with missing joints.}
    \label{fig:missingness-feature}
  \end{minipage}
\end{figure}

\paragraph{Temporal Parameterisation.}
The conditional distributions above are parameterised by recurrent networks that encode their stated
latent histories. The group network encodes \((z_{<t}^g,Z_{<t}^{\mathrm{ind}})\), while the person network
encodes \(z_{<t}^p\); both also receive the corresponding context representations from
\cref{ssec:context-enc}. Their LSTM activations are deterministic implementation details, rather than
additional variables in the probabilistic model, so the factorisations are written directly in terms of
latent histories and observed sets.

\paragraph{Precision-Weighted Posterior.}
Rather than parametrising the posterior directly, we compute it as a precision-weighted combination of two
signals: a \emph{bottom-up} proposal driven by the full target data and a \emph{top-down} term from the
prior, based on prior work by \citet{Snderby2016}. For the group latent, let
\(H_t^g=(z_{<t}^g,Z_{<t}^{\mathrm{ind}})\), and define
\((\mu_{\mathrm{bu}}, \sigma_{\mathrm{bu}}) = f_\phi^{\mathrm{bu}}(H_t^g,C,D)\) and
\((\mu_{\mathrm{td}}, \sigma_{\mathrm{td}}) = f_\theta^{\mathrm{td}}(H_t^g,C_t)\),
where \(f^{\mathrm{bu}}\) and \(f^{\mathrm{td}}\) are the bottom-up and top-down encoder networks
respectively. The fused posterior parameters are
obtained by precision merging:
\begin{equation}
  \sigma_q^{-2} = \sigma_{\mathrm{bu}}^{-2} + \sigma_{\mathrm{td}}^{-2},
  \qquad
  \mu_q = \sigma_q^2 \bigl(\sigma_{\mathrm{bu}}^{-2}\,\mu_{\mathrm{bu}} +
  \sigma_{\mathrm{td}}^{-2}\,\mu_{\mathrm{td}}\bigr).
  \label{eqn:precision-merge}
\end{equation}
This can be viewed as a product-of-experts combination~\citep{Hinton2002} of two diagonal Gaussians,
where each expert's influence is proportional to its precision. The same merging scheme is applied for each
person-level latent \(z_t^p\): the corresponding networks condition on
\((z_{<t}^p,z_t^g,C,D)\) and \((z_{<t}^p,z_t^g,C_t)\), respectively. Along with sharing information from
the generative and inference paths, this
precision-weighted formulation allows the posterior to smoothly interpolate between data-driven evidence and
the learned prior, which is
particularly beneficial when context is sparse or absent.

\paragraph{Training Objective.}
Combining the generative model in \cref{eqn:soc-gen-model} with the approximate posterior in
\cref{eqn:soc-approx-posterior} yields the following weighted variational objective:
{\small
  \begin{align}
    \mathrm{ELBO}_{\beta} &:= \sum_{t=1}^T
    \Bigl\langle \ln p_\theta(Y_t \mid X_t, z_t^g, Z_t^{\mathrm{ind}}) \Bigr\rangle_{q_\phi} \nonumber \\
    &\quad - \beta_g \sum_{t=1}^T
    \Bigl\langle\mathbb{KL}\bigl(q_\phi(z_t^g \mid z_{<t}^g,Z_{<t}^{\mathrm{ind}}, C, D) \;\|\;
    p_\theta(z_t^g \mid z_{<t}^g,Z_{<t}^{\mathrm{ind}}, C_t)\bigr)\Bigr\rangle_{q_\phi} \nonumber \\
    &\quad - \beta_p \sum_{t=1}^T \sum_{p=1}^P
    \Bigl\langle\mathbb{KL}\bigl(q_\phi(z_t^p \mid z_{<t}^p, z_t^g, C, D) \;\|\; p_\theta(z_t^p \mid
    z_{<t}^p, z_t^g, C_t)\bigr)\Bigr\rangle_{q_\phi},
    \label{eqn:soc-elbo}
\end{align}}%
where expectations include latent histories and, for person KLs, the current group latent.
For \(\beta_g=\beta_p=1\), this is a lower bound on \(\ln p_\theta(Y\mid X,C)\);
\Cref{appx:elbo-derivation} gives the derivation and weighting details.
We maximise \cref{eqn:soc-elbo}; equivalently, we minimise \(\mathcal{L}_{\mathrm{ELBO}} \coloneqq
-\mathrm{ELBO}_{\beta}\). We also supervise joint rotations, root-relative joint positions, and canonical
transforms with reconstruction losses \(\mathcal{L}_{\mathrm{pose}}\), \(\mathcal{L}_{\mathrm{key}}\), and
\(\mathcal{L}_{\mathrm{root}}\), respectively. The full training objective is
\begin{equation}
  \mathcal{L}_{\mathrm{train}}
  = \mathcal{L}_{\mathrm{ELBO}}
  + \lambda_{\mathrm{pose}} \mathcal{L}_{\mathrm{pose}}
  + \lambda_{\mathrm{key}} \mathcal{L}_{\mathrm{key}}
  + \lambda_{\mathrm{root}} \mathcal{L}_{\mathrm{root}}.
  \label{eqn:train-loss}
\end{equation}
Reconstruction-loss definitions and further training details are given in \Cref{appx:training-details}.

\section{Experiments}\label{sec:experiments}

\paragraph{Datasets.}
To evaluate the performance of \methodname, we conduct a number of
experiments using the
Haggling~\citep{Joo2016}, DnD~\citep{Mughal2024},
DD100~\citep{Siyao2024},
DuoBox~\citep{Cen2025} and Embody3D~\citep{McLean2025} datasets.
Unless otherwise stated, models have approximately 16M parameters and use AdamW~\citep{Loshchilov2017}
with a learning rate of \(1e-4\). Training details and model-size comparisons are given in
\Cref{appx:training-details,appx:model-size}.

\paragraph{Metrics.}
Multiparty generation requires more than agreement with a single recorded future: valid continuations can
differ in timing, role assignment, and interaction style. Standard reference errors (MPJPE, MPJVE) penalise
such alternatives without separating them from socially incoherent motion, so low error alone does not
establish the quality or variety of generated interactions. Likewise, diversity (DIV) is only meaningful
alongside distributional fit (Fr\'echet distance, FD), and foot skating (FS) and interpenetration (IP) flag
specific artefacts without establishing physical plausibility. We report these for comparability with prior
work, but prioritise interaction-level structure following~\citet{Shirekar2025}: synchrony, via
cross-recurrence quantification analysis as deviations in recurrence rate (\(\Delta\)RR) and determinism
(\(\Delta\)DET) from the ground truth, and structural similarity, via Soft-DTW (SDTW) and its cross-person
variant (Cr.\ SDTW), which compare trajectories up to natural timing variation. All metrics are
lower-is-better except DIV; \Cref{appx:evaluation-details} gives definitions and interpretation.

% \subsection{Social Forecasting}\label{ssec:social-forecasting}

% \textbf{Takeaway: \methodname remains competitive with state-of-the-art social forecasting while using a
% single model across dyadic, triadic, and five-person scenes.}
% \Cref{tab:social-forecasting-results} shows that \methodname obtains the lowest FD for every tested group size
% and the best MPJPE for larger groups, suggesting that the hierarchical latent state preserves coherent motion as
% the number of participants grows. Although random sampling can produce higher raw diversity, \methodname offers a
% better balance between realism, coordination, and reconstruction accuracy.

\subsection{Results}\label{ssec:results}

\textbf{A shared social-state formulation supports both forecasting future behaviour and completing missing
observations.}
\Cref{tab:social-forecasting-results,tab:tracking-infill-panoptic-dd100,tab:partner-dyadic-motion-prediction}
evaluate the reusable representation motivated in \Cref{sec:intro} through forecasting, inpainting, and
partner/dyadic prediction. The same group and person
latent formulation handles each setting by treating observed joints and people as context and unobserved
ones as targets. We first examine what social information the latent state retains, then assess how
this shared representation supports generation across tasks and observation regimes.

% \subsection{Hierarchical Latent Space Analysis}\label{ssec:hierarchical-reps}

\begin{figure}[!t]
  \centering
  \begin{minipage}[t]{0.42\textwidth}
    \vspace{0pt}
    \centering
    \includegraphics[width=\textwidth]{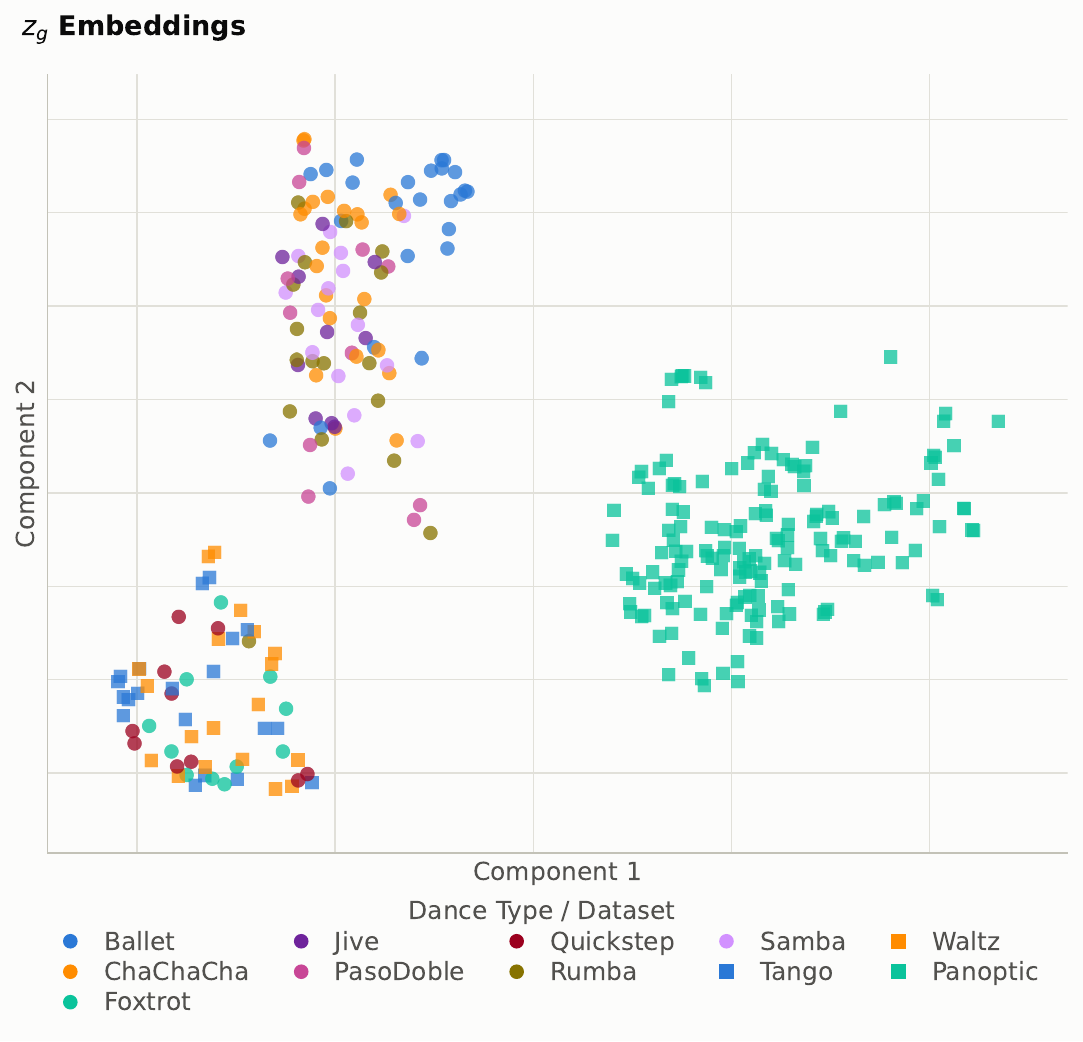}
    \caption{\(Z^g\) PACMAP~\citep{Wang2020} embeddings over DD100, and Panoptic Haggling.}
    \label{fig:zg-emb}
  \end{minipage}
  \hfill
  \begin{minipage}[t]{0.57\textwidth}
    \vspace{0pt}
    \centering
    \includegraphics[width=\textwidth]{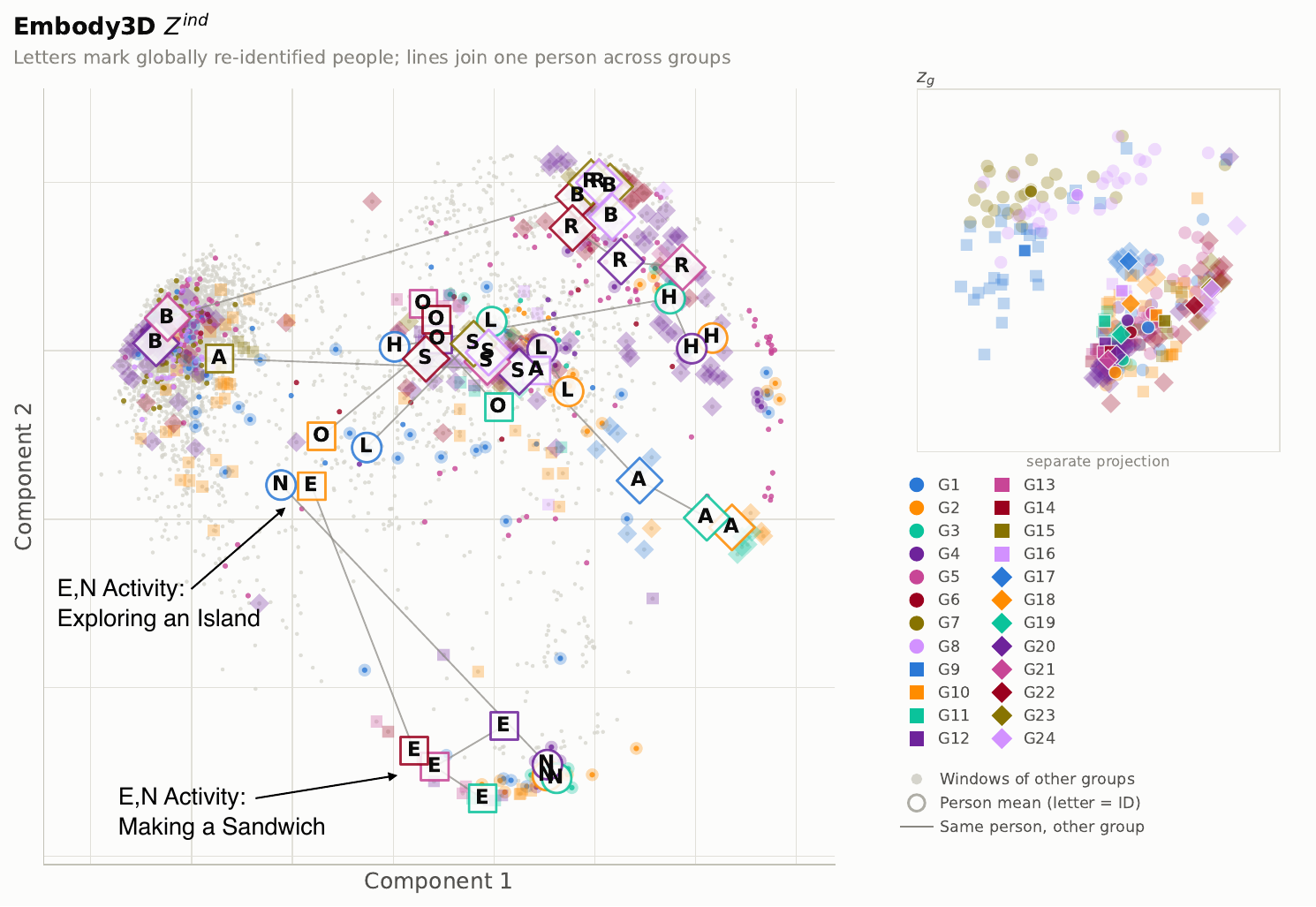}
    \caption{\(Z^{\text{ind}}\) PCA embeddings of Embody3D with common individuals across groups.
    Embeddings show the model latent space organising itself to cluster individuals.}
    \label{fig:zi-emb}
  \end{minipage}
\end{figure}

\textbf{Group latents organise activity structure, while person latents retain participant information across
changing groups.}
\Cref{fig:zg-emb} shows that the group latent \(Z^g\) forms structured clusters across DD100, Panoptic
Haggling, and within DD100 it separates dances based on whether they belong to the Latin or Western ballroom
dance styles. In
contrast, \Cref{fig:zi-emb} shows that \(Z^{\text{ind}}\) preserves person-specific structure: embeddings of the
same participant remain close even across different Embody3D groups. In the full latent space,
same-person distances have a median of \(6.36\), versus \(16.22\) for different-person pairs
(ratio \(0.392\)); the median same-person pair lies at the \(15.7\)th percentile of the
different-person distance distribution. Nearest-neighbour retrieval, excluding candidates from the
query's group, recovers the same individual in \(99/239\) queries (\(41.4\%\));
\Cref{appx:z-ind-analysis} gives the protocol and detailed results. This persistence across groups supports
the social-state aim in
\Cref{sec:intro}: retaining information about who is participating as the interaction context changes.
Together with the activity structure in \(Z^g\), it provides evidence that the latents carry inspectable
social information alongside their use for generation. Hierarchy ablations and qualitative probes in
\Cref{sec:hierarchy-ablation} further connect these representations to generation: \textit{the full hierarchy
  improves cross-person alignment over either single-level variant, while varying person latents or
changing the group-latent source produces different interaction realisations.}

% Measure the tables so body text wraps only beside the taller left table.
\newsavebox{\forecasttablebox}
\newsavebox{\inpaintingtablebox}
\newcount\resultwraplines
\begingroup
\sbox{\forecasttablebox}{%
  \begin{minipage}[t]{0.49\textwidth}
    \vspace{0pt}
    \centering
    \captionsetup{type=table}
    \caption{Social forecasting by observability and group size \(N\).
      Bold: best; underlined: second best per setting, excluding RS and NN.
      Models trained without the motion VAE (\Cref{appx:model-implementation}). \(^{\dagger}\)RS and NN
    ranked, as no learned baseline is available.}
    \label{tab:social-forecasting-results}
    \scriptsize
    \setlength{\tabcolsep}{2pt}
    \renewcommand{\arraystretch}{1.04}
    \begin{adjustbox}{max width=\linewidth}
      \begin{tabular}{@{}c l *{10}{c}@{}}
        \toprule
        \textbf{\(N\)} & \textbf{Method} & \textbf{\(\Delta\)RR\(\downarrow\)} & \textbf{\(\Delta\)DET\(\downarrow\)} &
        \textbf{SDTW\(\downarrow\)} & \textbf{Cr. SDTW\(\downarrow\)} & \textbf{IP\(\downarrow\)} &
        \textbf{FD\(\downarrow\)} & \textbf{DIV\(\uparrow\)} &
        \textbf{FS\(\downarrow\)} &
        \textbf{MPJPE\(\downarrow\)} & \textbf{MPJVE\(\downarrow\)} \\
        \midrule
        \multicolumn{12}{@{}l}{\textit{Full Observability}}\\
        2 & RS & 0.013 & 0.131 & 8.741 & 2.028 & <0.005 & 5.485 & 90.58 & 0.011 & 1.471 & 0.013 \\
        & NN & 0.013 & 0.093 & 8.292 & 2.059 & <0.005 & 0.319 & -- & <0.001 & 0.898 & 0.014 \\
        & MAGNet & \textbf{0.012} & \underline{0.350} & \textbf{6.041} & \textbf{1.348} &
        \underline{0.024} & \underline{0.490} & \underline{0.524} & \underline{0.570} & \underline{0.595} &
        \underline{0.185} \\
        & \textbf{\methodname} & \textbf{0.012} & \textbf{0.168} & \underline{7.416} & \underline{1.620} &
        \textbf{<0.005} & \textbf{0.239} & \textbf{2.316} & \textbf{0.001} & \textbf{0.305} &
        \textbf{0.013} \\
        \cmidrule{1-12}
        3 & RS & 0.013 & 0.166 & 8.533 & 1.898 & 0.222 & 2.131 & 80.86 & 0.012 & 1.647 & 0.010 \\
        & NN & 0.013 & 0.149 & 8.342 & 2.194 & 0.177 & 1.808 & -- & 0.018 & 1.120 & 0.009 \\
        & MAGNet & \underline{0.013} & \underline{0.335} & \textbf{7.318} & \textbf{1.531} &
        \underline{0.189} & \underline{0.378} & \underline{0.586} & \underline{0.452} & \underline{0.755} &
        \underline{0.437} \\
        & \textbf{\methodname} & \textbf{0.012} & \textbf{0.202} & \underline{8.025} & \underline{1.964} &
        \textbf{0.005} & \textbf{0.221} & \textbf{1.974} & \textbf{0.071} & \textbf{0.260} &
        \textbf{0.012} \\
        \cmidrule{1-12}
        4 & RS & 0.013 & 0.22 & 8.184 & 2.470 & 0.354 & 6.01 & 65.94 & 0.01 & 1.63 & 0.01 \\
        & NN & 0.013 & 0.18 & 8.324 & 2.194 & 0.177 & 1.80 & -- & 0.013 & 1.47 & 0.01 \\
        & MAGNet & \underline{0.013} & \underline{0.391} & \textbf{6.996} & \textbf{1.731} &
        \underline{0.232} & \underline{0.768} & \underline{0.710} & \underline{0.481} & \underline{0.937} &
        \underline{0.624} \\
        & \textbf{\methodname} & \textbf{0.012} & \textbf{0.197} & \underline{7.771} & \underline{1.919} &
        \textbf{0.022} & \textbf{0.216} & \textbf{3.19} & \textbf{0.099} & \textbf{0.353} &
        \textbf{0.015} \\
        \cmidrule{1-12}
        \(5^{\dagger}\) & RS & \underline{0.0128} & \underline{0.174} & \underline{8.279} & \textbf{1.542} &
        \(<0.005\) & \underline{1.348} & \textbf{21.61} & \textbf{0.0} & 0.751 & \underline{0.011} \\
        & NN & 0.0129 & \textbf{0.160} & \textbf{8.160} & \underline{1.549} & \underline{0.001} & \textbf{1.291} &
        -- & \textbf{0.0} & \underline{0.742} & \underline{0.011} \\
        & \textbf{\methodname} & \textbf{0.012} & 0.228 & 8.725 & 2.186 & \textbf{<0.001} & 1.766 &
        \underline{0.039} & \textbf{0.0} & \textbf{0.217} & \textbf{0.008} \\
        \midrule[1.2pt]
        \multicolumn{12}{@{}l}{\textit{Partial Observability}}\\
        2 & RoHM & \underline{0.013} & \underline{0.241} & \textbf{4.552} & \textbf{0.883} &
        \underline{0.147} & \underline{0.406} & \underline{0.093} & \underline{0.167} & \underline{0.489} &
        \underline{0.043} \\
        & \textbf{\methodname} & \textbf{0.012} & \textbf{0.168} & \underline{7.420} & \underline{1.60} &
        \textbf{0.0013} & \textbf{0.126} & \textbf{2.51} & \textbf{0.0012} & \textbf{0.400} &
        \textbf{0.016} \\
        \cmidrule{1-12}
        3 & RoHM & \textbf{0.013} & \underline{0.314} & \textbf{4.666} & \textbf{0.848} & \underline{0.168} &
        \underline{0.631} & \underline{0.081} & \textbf{0.173} & \underline{0.444} & \underline{0.041} \\
        & \textbf{\methodname} & \textbf{0.013} & \textbf{0.200} & \underline{8.06} & \underline{1.934} &
        \textbf{0.003} & \textbf{0.141} & \textbf{2.398} & \underline{0.235} & \textbf{0.297} &
        \textbf{0.015} \\
        \cmidrule{1-12}
        4 & RoHM & \underline{0.013} & \underline{0.302} & \textbf{4.593} & \textbf{0.862} &
        \underline{0.125} & \underline{0.701} & \underline{0.101} & \underline{0.186} & \underline{0.461} &
        \underline{0.041} \\
        & \textbf{\methodname} & \textbf{0.012} & \textbf{0.2} & \underline{7.80} & \underline{1.89} &
        \textbf{0.016} & \textbf{0.20} & \textbf{3.59} & \textbf{0.096} & \textbf{0.33} & \textbf{0.015} \\
        \cmidrule{1-12}
        5 & RoHM & \underline{0.014} & \textbf{0.206} & \textbf{8.130} & \textbf{1.019} &
        \underline{0.3626} & \textbf{0.32} & \textbf{0.107} & \underline{0.131} & \underline{0.533} &
        \underline{0.028} \\
        & \textbf{\methodname} & \textbf{0.012} & \underline{0.226} & \underline{8.722} &
        \underline{2.164} & \textbf{0.0005} & \underline{1.73} & \underline{0.039} & \textbf{0} &
        \textbf{0.213} & \textbf{0.008} \\
        \bottomrule
      \end{tabular}
    \end{adjustbox}
  \end{minipage}%
}
\sbox{\inpaintingtablebox}{%
  \begin{minipage}[t]{0.49\textwidth}
    \vspace{0pt}
    \centering
    \captionsetup{type=table}
    \caption{Partner inpainting on Panoptic and DD100.
    Bold: best; underlined: second best per setting, excluding RS and NN.}
    \label{tab:tracking-infill-panoptic-dd100}
    \scriptsize
    \setlength{\tabcolsep}{2pt}
    \renewcommand{\arraystretch}{1.04}
    \begin{adjustbox}{max width=\linewidth}
      \begin{tabular}{@{}l l *{10}{c}@{}}
        \toprule
        & \textbf{Method} & \textbf{\(\Delta\)RR\(\downarrow\)} & \textbf{\(\Delta\)DET\(\downarrow\)} &
        \textbf{SDTW\(\downarrow\)} & \textbf{Cr. SDTW\(\downarrow\)} & \textbf{IP\(\downarrow\)} &
        \textbf{FD\(\downarrow\)} &
        \textbf{DIV\(\uparrow\)} & \textbf{FS\(\downarrow\)} &
        \textbf{MPJPE\(\downarrow\)} &
        \textbf{MPJVE\(\downarrow\)} \\
        \midrule
        Panoptic & RS & 0.013 & 0.142 & 8.069 & 2.399 & 0.167 & 22.692 & 68.505 & 0.462 & 1.371 & 0.017 \\
        & NN & 0.013 & 0.129 & 8.483 & 1.932 & 0 & 4.203 & -- & 0.587 & 1.080 & 0.010 \\
        & Social Processes~\citeyearpar{Raman2023xgd} & \textbf{0.009} & \textbf{0.156} &
        \underline{8.952} & \underline{3.689} & \textbf{0} & \underline{0.951} & \underline{0.120} &
        \underline{0.385} & \underline{0.247} & \textbf{0.006} \\
        & \textbf{\methodname} & \textbf{0.009} & \underline{0.196} & \textbf{7.490} & \textbf{1.260} &
        \textbf{0} & \textbf{0.139} & \textbf{1.525} & \textbf{<0.005} & \textbf{0.217} &
        \underline{0.009} \\
        \cdashline{1-12}
        \multicolumn{12}{@{}l}{\textit{Partial Observability}}\\
        & RoHM~\citeyearpar{Zhang20244ao} & \underline{0.013} & \underline{0.221} & \underline{8.252} &
        \textbf{1.008} & \underline{0.179} & \underline{0.255} & \underline{0.030} & \underline{0.172} &
        \underline{0.417} & \underline{0.022} \\
        & \textbf{\methodname} & \textbf{0.009} & \textbf{0.195} & \textbf{7.493} & \underline{1.262} &
        \textbf{0.0} & \textbf{0.130} & \textbf{3.640} & \textbf{0.016} & \textbf{0.239} & \textbf{0.018} \\
        \cmidrule{1-12}
        DD100 & RS & 0.010 & 0.063 & 8.595 & 3.377 & 0.175 & 11.600 & 88.595 & 0.228 & 1.761 & 0.042 \\
        & NN & 0.009 & 0.044 & 7.673 & 2.018 & 0.302 & 1.420 & -- & 0.021 & 1.642 & 0.049 \\
        & Duolando~\citeyearpar{Siyao2024} & -- & -- & -- & -- & -- & 18.180 & 0.000 & 1.880 & 1.680 &
        0.070 \\
        & MAGNet~\citeyearpar{Maluleke2025} & 0.013 & 0.184 & \underline{5.646} & 2.022 &
        \underline{0.169} & \underline{0.42} & 0.71 & 0.873 & 1.187 & 0.036 \\
        & Social Processes~\citeyearpar{Raman2023xgd} & \textbf{0.007} & \underline{0.180} & 8.328 &
        \underline{1.937} & 0.332 & 2.05 & \underline{1.42} & \underline{0.097} & \underline{1.062} &
        \underline{0.033} \\
        & \textbf{\methodname} & \underline{0.008} & \textbf{0.06} & \textbf{4.590} & \textbf{1.527} &
        \textbf{0.117} & \textbf{0.075} & \textbf{7.72} & \textbf{0.012} & \textbf{1.050} &
        \textbf{0.028} \\
        \cdashline{1-12}
        \multicolumn{12}{@{}l}{\textit{Partial Observability}}\\
        & RoHM~\citeyearpar{Zhang20244ao} & \underline{0.013} & \underline{0.289} & \underline{7.017} &
        \textbf{1.494} & \underline{0.241} & \underline{0.173} & \underline{0.08} & \underline{0.311} &
        \underline{1.053} & \underline{0.047} \\
        & \textbf{\methodname} & \textbf{0.008} & \textbf{0.066} & \textbf{4.623} & \underline{1.514} &
        \textbf{0.118} & \textbf{0.075} & \textbf{7.49} & \textbf{0.013} & \textbf{1.050} &
        \textbf{0.028} \\
        \bottomrule
      \end{tabular}
    \end{adjustbox}
  \end{minipage}%
}
% Reserve the taller table's height and let the paragraph fill the space on its right.
\Needspace{\dimexpr\ht\forecasttablebox+\dp\forecasttablebox+\baselineskip\relax}
\noindent
\raisebox{0pt}[\ht\inpaintingtablebox][\dp\inpaintingtablebox]{\usebox{\forecasttablebox}}%
\hfill\usebox{\inpaintingtablebox}\par\smallskip
% Narrow only the lines beside the portion of Table 2 extending below Table 3.
\dimen0=\dimexpr\ht\forecasttablebox+\dp\forecasttablebox
-\ht\inpaintingtablebox-\dp\inpaintingtablebox+\baselineskip-1sp\relax
\divide\dimen0 by \baselineskip
\resultwraplines=\dimen0
% The Haggling dataset consists of
% triads playing a game where two people are sellers competing to
% sell their product to a buyer.  On the other hand,
% the DnD dataset consists of five people playing Dungeons and
% Dragons (DnD), a table top game. DnD typically
% has one member of the group acting as the ``dungeon master'' and
% the rest role play characters of their choice.
% Instead of using the raw keypoints from the data we first fit
% SMPL-X~\citep{Pavlakos2019} parameters on
% DnD and Haggling datasets to bring all datasets under a single
% unified representation.

\noindent\hangindent=0.51\textwidth\hangafter=-\resultwraplines
\textbf{Under partial observation, forecasting retains sample diversity and temporal structure while
improving distributional fit.}
\emph{Partial observability} here denotes joints missing through self-occlusion~\citep{Sigal2006-ec,Yao2022-lh}
or occlusion by other participants~\citep{Joo2016,Raman2022-st}, simulated by masking context joints as in
prior work~\citep{Zhang20244ao}; occluding all of a person's joints extends this to a missing
person.
Comparing \methodname's full and partial observation results in \Cref{tab:social-forecasting-results}, FD
decreases at every group size, including \(0.239\) to \(0.126\) for two people and \(0.221\) to \(0.141\)
for three. DIV increases for two to four people and is unchanged for five; SDTW changes by less than
\(0.5\%\), and cross-person SDTW improves throughout. This stability supports the introductory goal of a
social state that sustains generation as available evidence changes. The robustness is metric-specific:
three-person FS rises from \(0.071\) to \(0.235\), and MPJPE increases for two and three people.
Recorded Embody3D motion has zero FS, so the nonzero FS for three and four people is an  artefact.
Its interpenetration rate is also low (\(0.0002\)--\(0.0007\)); \methodname's partial-observation IP
(\(0.0013\)--\(0.016\)) remains above this level, though \(8\)--\(110\) times lower than RoHM's.
Against baselines, \methodname has the lowest FD for two to four people within each observation regime;
its lowest MPJPE at every group size further supports reference recovery.
\par
\endgroup

For five-person DnD, participants mostly stand around a table playing the game. \methodname's zero FS
matches the recorded motion, which also has zero FS, and its IP (at most \(0.001\)) stays below the
recorded rate of \(0.0016\). DIV \(0.039\) should be interpreted in this activity context, as the dataset
involves five people standing still quite often.
FD favours NN under full observations and RoHM under partial observations, identifying distributional matching
as a direction for improvement in this setting. Activity and group size vary together, precluding a
group-size-only explanation.

\textbf{Partner inpainting combines greater sample variation with better temporal alignment and less foot
skating than RoHM.}
Under partial observations, \methodname improves DIV, FS, FD, and SDTW over RoHM on both inpainting datasets
(\Cref{tab:tracking-infill-panoptic-dd100}). On DD100, DIV rises from
\(0.080\) to \(7.49\), while FS falls from \(0.311\) to \(0.013\) and SDTW from \(7.017\) to \(4.623\).
Recorded motion has zero FS on both datasets, so \methodname's FS is also closer to the data.
Interpenetration must be read against the recorded rate. Panoptic participants never touch, and
\methodname likewise produces no interpenetration, whereas RoHM reaches \(0.179\). DD100 dancers in close
hold overlap in \(29\%\) of recorded frames (IP \(0.288\)), so \methodname's lower IP (\(0.118\) versus
\(0.241\) for RoHM) reflects looser partner coupling rather than fewer artefacts.
Lower MPJPE additionally supports reference recovery, while RoHM's lower
cross-person SDTW on both datasets identifies closer interpersonal alignment as its complementary strength.

\textbf{A social state can represent varied futures while preserving distributional fit under partial
observation.}
Relative to RoHM, \methodname produces more diverse samples with lower FD in both DuoBox prediction settings
under partial observations (\Cref{tab:partner-dyadic-motion-prediction}). In dyadic prediction, DIV rises
from \(0.026\) to \(2.37\) and FD falls from \(0.325\) to \(0.115\). Foot skating must be read against
the data: fast boxing footwork gives the recorded motion an FS of \(0.276\) (\(0.283\) in partner
prediction). RoHM exceeds this rate (\(0.638\)), whereas \methodname falls below it (\(0.157\) and
\(0.172\)), consistent with generated joint speeds of roughly half the recorded ones. Dyadic IP
(\(0.009\)) remains about ten times the recorded rate of \(0.0008\). RoHM tracks the recorded future more
closely by MPJPE (\(0.660\) versus \(0.773\)).
For the social-state motivation, this supports representing possible continuations beyond a single recorded
trajectory. Such samples could supply candidate futures to a agent's planner.

\textbf{The social state carries structure from context, even across domains.}
A model trained only on DD100 dance, never shown conversation, still keeps a plausible conversational
formation when given \(40\) frames of held-out Panoptic context. What it lacks is the activity's own
dynamics: it drifts toward dance-like rotations, whereas adding conversational data lowers cross-person SDTW
from 1.424 to 0.511 and FD from 1.94 to 0.27 (\Cref{appx:cross-domain}).

% \subsection{Response Generation}

% In the Response Generation task, the goal is to produce appropriate
% behavioural or communicative responses
% for a target individual, conditioned on the observed actions and
% cues of others in the scene. The available
% context consists of recent interaction history, which provides
% information about both the target and
% surrounding participants. As in the other regimes, the task is
% inherently probabilistic, since similar
% interaction histories can lead to different plausible responses. We
% compare \methodname with [baseline
% methods], evaluating their ability to generate temporally coherent
% and socially appropriate responses over a
% rollout length of \(T=200\).

\begin{table}[!t]
  \centering
  \caption{Partner and dyadic forecasting with a missing partner on DuoBox.
  Bold: best; underlined: second best per setting, excluding RS and NN.}
  \label{tab:partner-dyadic-motion-prediction}
  \scriptsize
  \setlength{\tabcolsep}{2.5pt}
  \renewcommand{\arraystretch}{1.08}
  \begin{adjustbox}{max width=\linewidth}
    \begin{tabular}{@{}l *{20}{c}@{}}
      \toprule
      \textbf{Method} & \multicolumn{10}{c}{\textbf{Partner Motion Prediction}} &
      \multicolumn{10}{c}{\textbf{Dyadic Motion Prediction}} \\
      \cmidrule(lr){2-11}\cmidrule(lr){12-21}
      & \textbf{\(\Delta\)RR\(\downarrow\)} & \textbf{\(\Delta\)DET\(\downarrow\)} & \textbf{SDTW\(\downarrow\)} &
      \textbf{Cr. SDTW\(\downarrow\)} & \textbf{IP\(\downarrow\)} &
      \textbf{FD\(\downarrow\)} & \textbf{DIV\(\uparrow\)} &
      \textbf{FS\(\downarrow\)} & \textbf{MPJPE\(\downarrow\)} &
      \textbf{MPJVE\(\downarrow\)} &
      \textbf{\(\Delta\)RR\(\downarrow\)} & \textbf{\(\Delta\)DET\(\downarrow\)} & \textbf{SDTW\(\downarrow\)} &
      \textbf{Cr. SDTW\(\downarrow\)} & \textbf{IP\(\downarrow\)} & \textbf{FD\(\downarrow\)}
      & \textbf{DIV\(\uparrow\)} &
      \textbf{FS\(\downarrow\)} & \textbf{MPJPE\(\downarrow\)} &
      \textbf{MPJVE\(\downarrow\)} \\
      \midrule
      RS & 0.010 & 0.064 & 7.702 & 2.554 & 0.001 & 0.310 & 15.9 & 0.049 & 1.071 & 0.041 & 0.011 & 0.066 &
      7.698 & 2.552 & 0.002 & 0.119 & 16.232 & 0.047 & 1.074 & 0.042 \\
      NN & 0.009 & 0.064 & 6.598 & 2.525 & 0.001 & 0.766 & -- & 0.026 & 0.880 & 0.040 & 0.009 & 0.065 &
      6.604 & 2.528 & 0.002 & 0.295 & -- & 0.026 & 0.895 & 0.041 \\
      R2R~\citeyearpar{Cen2025} & -- & -- & -- & -- & -- & \underline{0.181} & \textbf{0.318} &
      \underline{0.255} & \underline{0.580} & \underline{0.029} & -- & -- & -- & -- & -- &
      \underline{0.337} & \textbf{0.395} & \underline{0.249} & \textbf{0.624} & \textbf{0.029} \\
      MAGNet~\citeyearpar{Maluleke2025} & 0.007 & 0.069 & 1.560 & 0.444 & 0.001 & \textbf{0.057} &
      \underline{0.034} & \textbf{0.071} & \textbf{0.125} & \textbf{0.022} & 0.012 & 0.120 & 4.564 &
      1.886 & 0.004 & \textbf{0.102} & \underline{0.370} & \textbf{0.087} & \underline{0.697} &
      \underline{0.032} \\
      \cdashline{1-21}
      \multicolumn{21}{@{}l}{\textit{Partial Observability}}\\
      RoHM~\citeyearpar{Zhang20244ao} & \underline{0.014} & \underline{0.166} & \textbf{3.457} &
      \underline{1.551} & \underline{0.095} & \underline{0.325} & \underline{0.026} & \underline{0.638} &
      \textbf{0.660} & \underline{0.041} & \underline{0.0134} & \underline{0.231} & \underline{6.914} &
      \underline{1.760} & \underline{0.210} & \underline{0.325} & \underline{0.026} & \underline{0.638} &
      \textbf{0.660} & \underline{0.041} \\
      \textbf{\methodname} & \textbf{0.007} & \textbf{0.087} & \underline{6.360} & \textbf{0.971} &
      \textbf{0.009} & \textbf{0.116} & \textbf{2.39} & \textbf{0.172} & \underline{0.912} &
      \textbf{0.030} & \textbf{0.008} & \textbf{0.081} & \textbf{5.743} & \textbf{0.945} & \textbf{0.009} &
      \textbf{0.115} & \textbf{2.37} & \textbf{0.157} & \underline{0.773} & \textbf{0.031} \\
      \bottomrule
    \end{tabular}
  \end{adjustbox}
\end{table}

\section{Discussion and Conclusion}\label{sec:discussion-conclusion}

\methodname frames multi-person motion as conditional generation over arbitrary context sets: a single
joint-level interface covers full or sparse observation, missing joints or people, and future rollout without
task-specific heads. Across forecasting, partner inpainting, and dyadic prediction, it retains distributional
fit and temporal alignment as evidence is removed, and produces varied completions rather than a single
averaged future. Its latent social state is also inspectable: without supervision, \(Z^g\) organises
interactions by activity and dance style, and \(Z^{\mathrm{ind}}\) recovers the same participant across
different groups. The model also generalises formation structure across domains
(\Cref{appx:cross-domain}).

These two levels of social state open several directions. Because \(Z^g\) summarises the ongoing interaction,
it could serve as a compact belief state for agents that observe or join a group, conditioning their policies
on the phase of an interaction or detecting when it shifts. Because \(Z^{\mathrm{ind}}\) persists across
groups, it could support memory of who someone is across encounters and adaptation to an individual's
interaction style. More broadly, both states could be supplied to foundation models, vision-language-action
models, and robot controllers as a compact summary of human interaction, helping them gauge how a group is
coordinated and how each person tends to behave without reasoning over raw pose streams. Editing or
exchanging the two levels, as in the group-latent switch of \Cref{sec:hierarchy-ablation}, suggests
controllable synthesis of
social scenarios for training and evaluating agents.
% \paragraph{Limitations.}
The evidence covers the evaluated social-motion datasets and depends on SMPL-X preprocessing quality.
Diversity is low in the largely stationary five-person DnD scenes, and pose and timing errors can accumulate
over long open-loop rollouts; since new observations can enter the context at any time, however, the model
can correct its predictions as evidence arrives, as in tracking. The latent analyses show useful structure
but not complete disentanglement; incorporating \methodname in a simulator or agent world model is the
natural next step.

\subsection*{AI Use Statement}

We have not used generative AI tools for proposing or refining hypotheses, generating synthetic data sets,
cleaning and reformatting datasets, and supporting qualitative and thematic data analysis
and providing critical ingredients for proving mathematical claims, assisting in the writing of proofs, and
assisting with translation are not applicable to this work.
We used generative AI tools for modifying scientific figures, creating or editing software code, and editing
the research paper to improve readability. We have reviewed all AI-assisted work.

\bibliographystyle{plainnat}
\bibliography{references,paperpile}

@ARTICLE{Singh2019-qw,
  title         = "Sequential Neural Processes",
  author        = "Singh, Gautam and Yoon, Jaesik and Son, Youngsung and Ahn,
                   Sungjin",
  journal       = "arXiv [cs.LG]",
  month         =  "24~" # jun,
  year          =  2019,
  url           = "http://arxiv.org/abs/1906.10264",
  archivePrefix = "arXiv",
  primaryClass  = "cs.LG"
}

@ARTICLE{Raman2022-st,
  title     = "{ConfLab}: A data collection concept, dataset, and benchmark for
               machine analysis of free-standing social interactions in the wild",
  author    = "Raman, Chirag and Vargas-Quiros, Jose and Tan, Stephanie and
               Islam, Ashraful and Gedik, Ekin and Hung, H",
  editor    = "Koyejo, S and Mohamed, S and Agarwal, A and Belgrave, D and Cho,
               K and Oh, A",
  journal   = "Neural Information Processing Systems",
  publisher = "Curran Associates, Inc.",
  volume    =  35,
  pages     = "23701--23715",
  month     =  "10~" # may,
  year      =  2022,
  url       = "https://proceedings.neurips.cc/paper_files/paper/2022/hash/95f9ad2e251e9014697589037450f9bb-Abstract-Datasets_and_Benchmarks.html"
}

@ARTICLE{Xu2024-xs,
  title         = "{ReGenNet}: Towards human action-reaction synthesis",
  author        = "Xu, Liang and Zhou, Yizhou and Yan, Yichao and Jin, Xin and
                   Zhu, Wenhan and Rao, Fengyun and Yang, Xiaokang and Zeng,
                   Wenjun",
  journal       = "arXiv [cs.CV]",
  pages         = "1759--1769",
  month         =  "18~" # mar,
  year          =  2024,
  url           = "https://openaccess.thecvf.com/content/CVPR2024/html/Xu_ReGenNet_Towards_Human_Action-Reaction_Synthesis_CVPR_2024_paper.html",
  archivePrefix = "arXiv",
  primaryClass  = "cs.CV",
  doi           = "10.48550/arXiv.2403.11882"
}

@INPROCEEDINGS{Sigal2006-ec,
  title     = "Measure locally, reason globally: Occlusion-sensitive articulated
               pose estimation",
  author    = "Sigal, L and Black, M J",
  booktitle = "2006 IEEE Computer Society Conference on Computer Vision and
               Pattern Recognition - Volume 2 (CVPR'06)",
  publisher = "IEEE",
  year      =  2006,
  url       = "https://ieeexplore.ieee.org/document/1641003",
  doi       = "10.1109/cvpr.2006.180",
  isbn      =  9780769525976,
  language  = "en"
}

@INCOLLECTION{Yao2022-lh,
  title     = "Learning visibility for robust dense human body estimation",
  author    = "Yao, Chun-Han and Yang, Jimei and Ceylan, Duygu and Zhou, Yi and
               Zhou, Yang and Yang, Ming-Hsuan",
  booktitle = "Lecture Notes in Computer Science",
  publisher = "Springer Nature Switzerland",
  address   = "Cham",
  pages     = "412--428",
  series    = "Lecture Notes in Computer Science",
  year      =  2022,
  url       = "http://dx.doi.org/10.1007/978-3-031-19769-7_24",
  doi       = "10.1007/978-3-031-19769-7\_24",
  isbn      = "9783031197680,9783031197697",
  language  = "en"
}

@article{Cen2025,
  year      = {2025},
  title     = {{Ready-to-React: Online Reaction Policy for Two-Character Interaction Generation}},
  author    = {Cen, Zhi and Pi, Huaijin and Peng, Sida and Shuai, Qing and Shen, Yujun and Bao, Hujun and Zhou, Xiaowei and Hu, Ruizhen},
  journal   = {arXiv},
  doi       = {10.48550/arxiv.2502.20370},
  eprint    = {2502.20370}
}

@article{Chartrand1999,
  year     = {1999},
  title    = {The chameleon effect: The perception–behavior link and social interaction},
  author   = {Chartrand, Tanya L and Bargh, John A},
  journal  = {Journal of personality and social psychology},
  issn     = {0022-3514},
  doi      = {10.1037/0022-3514.76.6.893},
  url      = {https://doi.apa.org/doi/10.1037/0022-3514.76.6.893},
  pages    = {893--910},
  number   = {6},
  volume   = {76}
}

@article{Chen2022o2e,
  year      = {2022},
  title     = {{Executing your Commands via Motion Diffusion in Latent Space}},
  author    = {Chen, Xin and Jiang, Biao and Liu, Wen and Huang, Zilong and Fu, Bin and Chen, Tao and Yu, Jingyi and Yu, Gang},
  journal   = {arXiv},
  doi       = {10.48550/arxiv.2212.04048},
  eprint    = {2212.04048}
}

@article{Chew2025,
  year     = {2025},
  title    = {{SBM: Social Behavior Model for Human-Like Action Generation}},
  author   = {Chew, Jouh Yeong and Lin, Zhi-Yi and Zhang, Xucong},
  journal  = {Companion Proceedings of the 27th International Conference on Multimodal Interaction},
  doi      = {10.1145/3747327.3763038},
  pages    = {32--36}
}

@article{Diomataris2024,
  year     = {2024},
  title    = {{WANDR}: Intention-guided Human Motion Generation},
  author   = {Diomataris, Markos and Athanasiou, Nikos and Taheri, Omid and Wang, Xi and Hilliges, Otmar and Black, Michael J},
  journal  = {{arXiv}},
  doi      = {10.48550/arxiv.2404.15383},
  eprint   = {2404.15383},
  url      = {http://arxiv.org/abs/2404.15383},
  month    = {4}
}

@article{Feynman1955,
  year    = {1955},
  title   = {Slow Electrons in a Polar Crystal},
  author  = {Feynman, R. P.},
  journal = {Physical Review},
  issn    = {0031-899X},
  doi     = {10.1103/physrev.97.660},
  pages   = {660--665},
  number  = {3},
  volume  = {97}
}

@book{Feynman1972,
  title     = {Statistical mechanics: a set of lectures},
  author    = {Feynman, R. P.},
  isbn      = {9788187169970},
  url       = {https://books.google.nl/books?id=Vflsca1UFZoC},
  series    = {Frontiers in physics : a lecture note and reprint series},
  publisher = {Sarat Book Distributors},
  year      = {1972}
}

@article{Garnelo2018,
  year     = {2018},
  title    = {{Neural Processes}},
  author   = {Garnelo, Marta and Schwarz, Jonathan and Rosenbaum, Dan and Viola, Fabio and Rezende, Danilo J and Eslami, S M Ali and Teh, Yee Whye},
  journal  = {arXiv},
  doi      = {10.48550/arxiv.1807.01622},
  eprint   = {1807.01622},
  url      = {http://arxiv.org/abs/1807.01622},
  month    = {7}
}

@article{Gregor2018,
  year      = {2018},
  title     = {{Temporal Difference Variational Auto-Encoder}},
  author    = {Gregor, Karol and Papamakarios, George and Besse, Frederic and Buesing, Lars and Weber, Theophane},
  journal   = {arXiv},
  doi       = {10.48550/arxiv.1806.03107},
  eprint    = {1806.03107}
}

@article{Guo2022,
  year      = {2022},
  title     = {{Generating Diverse and Natural 3D Human Motions from Text}},
  author    = {Guo, Chuan and Zou, Shihao and Zuo, Xinxin and Wang, Sen and Ji, Wei and Li, Xingyu and Cheng, Li},
  journal   = {2022 IEEE/CVF Conference on Computer Vision and Pattern Recognition (CVPR)},
  doi       = {10.1109/cvpr52688.2022.00509},
  pages     = {5142--5151},
  volume    = {00}
}

@article{Hale2020,
  year     = {2020},
  title    = {Are you on my wavelength? Interpersonal coordination in dyadic       conversations},
  author   = {Hale, Joanna and Ward, Jamie A and Buccheri, Francesco and Oliver, Dominic and Hamilton, Antonia F de C},
  journal  = {Journal of nonverbal behavior},
  issn     = {0191-5886},
  doi      = {10.1007/s10919-019-00320-3},
  url      = {http://dx.doi.org/10.1007/s10919-019-00320-3},
  pages    = {63--83},
  number   = {1},
  volume   = {44}
}

@article{Hinton2002,
  year      = {2002},
  title     = {{Training Products of Experts by Minimizing Contrastive Divergence}},
  author    = {Hinton, Geoffrey E.},
  journal   = {Neural Computation},
  issn      = {0899-7667},
  doi       = {10.1162/089976602760128018},
  pmid      = {12180402},
  pages     = {1771--1800},
  number    = {8},
  volume    = {14}
}

@article{Hoehl2020,
  year    = {2020},
  title   = {Interactional synchrony: signals, mechanisms and benefits},
  author  = {Hoehl, Stefanie and Fairhurst, Merle and Schirmer, Annett},
  journal = {Social Cognitive and Affective Neuroscience},
  issn    = {1749-5016},
  doi     = {10.1093/scan/nsaa024},
  pmid    = {32128587},
  pmcid   = {{PMC}7812629},
  pages   = {5--18},
  number  = {1-2},
  volume  = {16}
}

@article{Jaegle2021,
  year     = {2021},
  title    = {{Perceiver: General Perception with Iterative Attention}},
  author   = {Jaegle, Andrew and Gimeno, Felix and Brock, Andrew and Zisserman, Andrew and Vinyals, Oriol and Carreira, Joao},
  journal  = {arXiv},
  doi      = {10.48550/arxiv.2103.03206},
  eprint   = {2103.03206}
}

@article{Joo2016,
  year     = {2016},
  title    = {Panoptic Studio: A Massively Multiview System for Social Interaction Capture},
  author   = {Joo, Hanbyul and Simon, Tomas and Li, Xulong and Liu, Hao and Tan, Lei and Gui, Lin and Banerjee, Sean and Godisart, Timothy and Nabbe, Bart and Matthews, Iain and Kanade, Takeo and Nobuhara, Shohei and Sheikh, Yaser},
  journal  = {{arXiv}},
  issn     = {1612.0315},
  doi      = {10.48550/arxiv.1612.03153},
  eprint   = {1612.03153},
  url      = {http://arxiv.org/abs/1612.03153},
  month    = {12}
}

@article{Kim2019l993,
  year      = {2019},
  title     = {{Attentive Neural Processes}},
  author    = {Kim, Hyunjik and Mnih, Andriy and Schwarz, Jonathan and Garnelo, Marta and Eslami, Ali and Rosenbaum, Dan and Vinyals, Oriol and Teh, Yee Whye},
  journal   = {arXiv},
  doi       = {10.48550/arxiv.1901.05761},
  eprint    = {1901.05761}
}

@article{Kingma2016,
  year      = {2016},
  title     = {{Improving Variational Inference with Inverse Autoregressive Flow}},
  author    = {Kingma, Diederik P and Salimans, Tim and Jozefowicz, Rafal and Chen, Xi and Sutskever, Ilya and Welling, Max},
  journal   = {arXiv},
  doi       = {10.48550/arxiv.1606.04934},
  eprint    = {1606.04934}
}

@article{Kleef2007,
  year     = {2007},
  title    = {Group member prototypicality and intergroup negotiation: How one's       standing in the group affects negotiation behaviour},
  author   = {Kleef, Gerben A Van and Steinel, Wolfgang and Knippenberg, Daan Van and Hogg, Michael A and Svensson, Alicia},
  journal  = {British Journal of Social Psychology},
  issn     = {0144-6665},
  doi      = {10.1348/014466605x89353},
  url      = {http://dx.doi.org/10.1348/014466605X89353},
  pages    = {129--152},
  number   = {1},
  volume   = {46},
  month    = {3}
}

@article{Levinson2015,
  year     = {2015},
  title    = {Timing in turn-taking and its implications for processing models of       language},
  author   = {Levinson, Stephen C and Torreira, Francisco},
  journal  = {Frontiers in psychology},
  issn     = {1664-1078},
  doi      = {10.3389/fpsyg.2015.00731},
  url      = {https://www.ncbi.nlm.nih.gov/pmc/articles/PMC4464110},
  pages    = {731},
  volume   = {6},
  month    = {6}
}

@article{Lin2026,
  year     = {2026},
  title    = {{PolySLGen: Online Multimodal Speaking-Listening Reaction Generation in Polyadic Interaction}},
  author   = {Lin, Zhi-Yi and Markhorst, Thomas and Chew, Jouh Yeong and Zhang, Xucong},
  journal  = {arXiv},
  doi      = {10.48550/arxiv.2604.08125},
  eprint   = {2604.08125}
}

@article{Maluleke2025,
  year    = {2025},
  title   = {Diffusion Forcing for Multi-Agent Interaction Sequence Modeling},
  author  = {Maluleke, Vongani H. and Horiuchi, Kie and Wilken, Lea and Ng, Evonne and Malik, Jitendra and Kanazawa, Angjoo},
  journal = {{arXiv}},
  doi     = {10.48550/arxiv.2512.17900},
  eprint  = {2512.17900}
}

@article{Markhorst2026,
  year     = {2026},
  title    = {{MuPPet: Multi-person 2D-to-3D Pose Lifting}},
  author   = {Markhorst, Thomas and Lin, Zhi-Yi and Chew, Jouh Yeong and Gemert, Jan van and Zhang, Xucong},
  journal  = {arXiv},
  doi      = {10.48550/arxiv.2604.09715},
  eprint   = {2604.09715}
}

@article{Matsumoto2007,
  year     = {2007},
  title    = {Culture, context, and behavior},
  author   = {Matsumoto, David},
  journal  = {Journal of personality},
  issn     = {1467-6494},
  doi      = {10.1111/j.1467-6494.2007.00476.x},
  url      = {https://www.ncbi.nlm.nih.gov/pubmed/17995466},
  pages    = {1285--1319},
  number   = {6},
  volume   = {75},
  month    = {12}
}

@article{McLean2025,
  year    = {2025},
  title   = {Embody 3D: A Large-scale Multimodal Motion and Behavior Dataset},
  author  = {{McLean}, Claire and Meendering, Makenzie and Swartz, Tristan and Gabbay, Orri and Olsen, Alexandra and Jacobs, Rachel and Rosen, Nicholas and Bree, Philippe de and Garcia, Tony and Merrill, Gadsden and Sandakly, Jake and Buffalini, Julia and Jain, Neham and Krenn, Steven and Kumar, Moneish and Markovic, Dejan and Ng, Evonne and Prada, Fabian and Saba, Andrew and Zhang, Siwei and Agrawal, Vasu and Godisart, Tim and Richard, Alexander and Zollhoefer, Michael},
  journal = {{arXiv}},
  doi     = {10.48550/arxiv.2510.16258},
  eprint  = {2510.16258}
}

@article{Mughal2024,
  year     = {2024},
  title    = {{ConvoFusion}: Multi-Modal Conversational Diffusion for Co-Speech Gesture Synthesis},
  author   = {Mughal, Muhammad Hamza and Dabral, Rishabh and Habibie, Ikhsanul and Donatelli, Lucia and Habermann, Marc and Theobalt, Christian},
  journal  = {{arXiv}},
  doi      = {10.48550/arxiv.2403.17936},
  eprint   = {2403.17936},
  url      = {http://arxiv.org/abs/2403.17936},
  month    = {3}
}

@article{Oers2005,
  year     = {2005},
  title    = {Context dependence of personalities: risk-taking behavior in a social and       a nonsocial situation},
  author   = {Oers, Kees van and Klunder, Margreet and Drent, Piet J},
  journal  = {Behavioral ecology: official journal of the International Society for Behavioral Ecology},
  issn     = {1045-2249},
  doi      = {10.1093/beheco/ari045},
  url      = {http://academic.oup.com/beheco/article/16/4/716/214729/Context-dependence-of-personalities-risktaking},
  pages    = {716--723},
  number   = {4},
  volume   = {16},
  month    = {7}
}

@article{Pandey2022,
  year      = {2022},
  title     = {{DiffuseVAE: Efficient, Controllable and High-Fidelity Generation from Low-Dimensional Latents}},
  author    = {Pandey, Kushagra and Mukherjee, Avideep and Rai, Piyush and Kumar, Abhishek},
  journal   = {arXiv},
  doi       = {10.48550/arxiv.2201.00308},
  eprint    = {2201.00308}
}

@article{Peterson1987,
  year    = {1987},
  title   = {A mean field theory learning algorithm for neural networks},
  author  = {Peterson, Carsten and Anderson, James R},
  journal = {Complex systems},
  number  = {5},
  volume  = {1}
}

@article{Preechakul2021,
  year      = {2021},
  title     = {{Diffusion Autoencoders: Toward a Meaningful and Decodable Representation}},
  author    = {Preechakul, Konpat and Chatthee, Nattanat and Wizadwongsa, Suttisak and Suwajanakorn, Supasorn},
  journal   = {arXiv},
  doi       = {10.48550/arxiv.2111.15640},
  eprint    = {2111.15640}
}

@article{Pruitt2011,
  year     = {2011},
  title    = {How within-group behavioural variation and task efficiency enhance fitness       in a social group},
  author   = {Pruitt, Jonathan N and Riechert, Susan E},
  journal  = {Proceedings. Biological sciences},
  issn     = {0962-8452},
  doi      = {10.1098/rspb.2010.1700},
  url      = {https://www.ncbi.nlm.nih.gov/pmc/articles/PMC3049074},
  pages    = {1209--1215},
  number   = {1709},
  volume   = {278},
  month    = {4}
}

@article{Ptschulat2026,
  year    = {2026},
  title   = {Transcript frame analysis: Thinking with Goffman about interview data},
  author  = {Pötschulat, Maike},
  journal = {Qualitative Research},
  issn    = {1468-7941},
  doi     = {10.1177/14687941251398982}
}

@incollection{Raman2023xgd,
  year      = {2023},
  title     = {Social Processes: Self-supervised Meta-learning Over Conversational Groups for Forecasting Nonverbal Social Cues},
  author    = {Raman, Chirag and Hung, Hayley and Loog, Marco},
  booktitle = {{ECCV} 2022 Workshops: Tel Aviv, Israel, October 23–27, 2022},
  isbn      = {9783031250651},
  url       = {https://link.springer.com/10.1007/978-3-031-25066-8_37},
  pages     = {639--659},
  series    = {Lecture Notes in Computer Science},
  publisher = {Springer Nature Switzerland},
  address   = {Cham},
  doi       = {10.1007/978-3-031-25066-8_37}
}

@article{Rathbone2023,
  year     = {2023},
  title    = {The reciprocal relationship between social identity and adherence to group       norms},
  author   = {Rathbone, Joanne A and Cruwys, Tegan and Stevens, Mark and Ferris, Laura J and Reynolds, Katherine J},
  journal  = {The British journal of social psychology},
  issn     = {2044-8309},
  doi      = {10.1111/bjso.12635},
  url      = {https://www.ncbi.nlm.nih.gov/pubmed/36786397},
  pages    = {1346--1362},
  number   = {3},
  volume   = {62},
  month    = {7}
}

@article{Sacks1974,
  year     = {1974},
  title    = {A Simplest Systematics for the Organization of Turn-Taking for Conversation},
  author   = {Sacks, Harvey and Schegloff, Emanuel A and Jefferson, Gail},
  journal  = {Language},
  issn     = {0097-8507},
  doi      = {10.2307/412243},
  url      = {https://www.jstor.org/stable/412243?origin=crossref},
  pages    = {696},
  number   = {4},
  volume   = {50},
  month    = {12}
}

@article{Shirekar2025,
  year     = {2025},
  title    = {{Multimodal Quantitative Measures for Multiparty Behavior Evaluation}},
  author   = {Shirekar, Ojas and Pouw, Wim and Hao, Chenxu and Phadnis, Vrushank and Beeler, Thabo and Raman, Chirag},
  journal  = {Proceedings of the 27th International Conference on Multimodal Interaction},
  doi      = {10.1145/3716553.3750752},
  eprint   = {2508.10916},
  pages    = {249--264},
  month    = {8}
}

@article{Siyao2024,
  year    = {2024},
  title   = {Duolando: Follower {GPT} with Off-Policy Reinforcement Learning for Dance Accompaniment},
  author  = {Siyao, Li and Gu, Tianpei and Yang, Zhitao and Lin, Zhengyu and Liu, Ziwei and Ding, Henghui and Yang, Lei and Loy, Chen Change},
  journal = {{arXiv}},
  doi     = {10.48550/arxiv.2403.18811},
  eprint  = {2403.18811}
}

@article{Smith2009,
  year     = {2009},
  title    = {Group Norms and the Attitude-Behaviour Relationship: Group norms and       attitude-behaviour relations},
  author   = {Smith, Joanne R and Louis, Winnifred R},
  journal  = {Social and personality psychology compass},
  issn     = {1751-9004},
  doi      = {10.1111/j.1751-9004.2008.00161.x},
  url      = {https://compass.onlinelibrary.wiley.com/doi/full/10.1111/j.1751-9004.2008.00161.x},
  pages    = {19--35},
  number   = {1},
  volume   = {3},
  month    = {1}
}

@article{Snderby2016,
  year    = {2016},
  title   = {Ladder Variational Autoencoders},
  author  = {Sønderby, Casper Kaae and Raiko, Tapani and Maaløe, Lars and Sønderby, Søren Kaae and Winther, Ole},
  journal = {{arXiv}},
  doi     = {10.48550/arxiv.1602.02282},
  eprint  = {1602.02282}
}

@article{Su2021,
  year      = {2021},
  title     = {{RoFormer: Enhanced Transformer with Rotary Position Embedding}},
  author    = {Su, Jianlin and Lu, Yu and Pan, Shengfeng and Murtadha, Ahmed and Wen, Bo and Liu, Yunfeng},
  journal   = {arXiv},
  doi       = {10.48550/arxiv.2104.09864},
  eprint    = {2104.09864}
}

@article{Tanke2023,
  year     = {2023},
  title    = {Social Diffusion: Long-term Multiple Human Motion Anticipation},
  author   = {Tanke, Julian and Zhang, Linguang and Zhao, Amy and Tang, Chengcheng and Cai, Yujun and Wang, Lezi and Wu, Po-Chen and Gall, Juergen and Keskin, Cem},
  journal  = {2023 {IEEE}/{CVF} International Conference on Computer Vision ({ICCV})},
  doi      = {10.1109/iccv51070.2023.00880},
  url      = {http://dx.doi.org/10.1109/ICCV51070.2023.00880},
  pages    = {9567--9577},
  volume   = {00},
  month    = {10}
}

@article{Terry1999,
  year     = {1999},
  title    = {The theory of planned behaviour: Self‐identity, social identity and group       norms},
  author   = {Terry, Deborah J and Hogg, Michael A and White, Katherine M},
  journal  = {The British journal of social psychology},
  issn     = {2044-8309},
  doi      = {10.1348/014466699164149},
  url      = {https://bpspsychub.onlinelibrary.wiley.com/doi/abs/10.1348/014466699164149},
  pages    = {225--244},
  number   = {3},
  volume   = {38},
  month    = {9}
}

@article{Tevet2022dzu,
  year      = {2022},
  title     = {{MotionCLIP: Exposing human motion generation to CLIP space}},
  author    = {Tevet, Guy and Gordon, Brian and Hertz, Amir and Bermano, Amit H and Cohen-Or, Daniel},
  journal   = {arXiv [cs.CV]},
  issn      = {2203.0806},
  url       = {http://arxiv.org/abs/2203.08063},
  month     = {3}
}

@article{Tevet2022h1j,
  year      = {2022},
  title     = {{Human Motion Diffusion Model}},
  author    = {Tevet, Guy and Raab, Sigal and Gordon, Brian and Shafir, Yonatan and Cohen-Or, Daniel and Bermano, Amit H},
  journal   = {arXiv},
  issn      = {2209.1491},
  doi       = {10.48550/arxiv.2209.14916},
  eprint    = {2209.14916},
  url       = {http://arxiv.org/abs/2209.14916},
  month     = {9}
}

@article{Vahdat2020,
  year    = {2020},
  title   = {{NVAE}: A Deep Hierarchical Variational Autoencoder},
  author  = {Vahdat, Arash and Kautz, Jan},
  journal = {{arXiv}},
  doi     = {10.48550/arxiv.2007.03898},
  eprint  = {2007.03898}
}

@article{Wang2020,
  year    = {2020},
  title   = {Understanding How Dimension Reduction Tools Work: An Empirical Approach to Deciphering t-{SNE}, {UMAP}, {TriMAP}, and {PaCMAP} for Data Visualization},
  author  = {Wang, Yingfan and Huang, Haiyang and Rudin, Cynthia and Shaposhnik, Yaron},
  journal = {{arXiv}},
  doi     = {10.48550/arxiv.2012.04456},
  eprint  = {2012.04456}
}

@article{Wang2021xhw,
  year      = {2021},
  title     = {{Multi-Person 3D Motion Prediction with Multi-Range Transformers}},
  author    = {Wang, Jiashun and Xu, Huazhe and Narasimhan, Medhini and Wang, Xiaolong},
  journal   = {arXiv},
  doi       = {10.48550/arxiv.2111.12073},
  eprint    = {2111.12073}
}

@article{Yi2024,
  year    = {2024},
  title   = {Estimating Body and Hand Motion in an Ego-sensed World},
  author  = {Yi, Brent and Ye, Vickie and Zheng, Maya and Li, Yunqi and Müller, Lea and Pavlakos, Georgios and Ma, Yi and Malik, Jitendra and Kanazawa, Angjoo},
  journal = {{arXiv}},
  doi     = {10.48550/arxiv.2410.03665},
  eprint  = {2410.03665}
}

@article{Zanlungo2017,
  year     = {2017},
  title    = {Intrinsic group behaviour: Dependence of pedestrian dyad dynamics on       principal social and personal features},
  author   = {Zanlungo, Francesco and Yücel, Zeynep and Brščić, Dražen and Kanda, Takayuki and Hagita, Norihiro},
  journal  = {{PloS} one},
  issn     = {1932-6203},
  doi      = {10.1371/journal.pone.0187253},
  url      = {https://www.ncbi.nlm.nih.gov/pmc/articles/PMC5667819},
  pages    = {e0187253},
  number   = {11},
  volume   = {12},
  month    = {11}
}

@article{Zhou2018,
  year    = {2018},
  title   = {On the Continuity of Rotation Representations in Neural Networks},
  author  = {Zhou, Yi and Barnes, Connelly and Lu, Jingwan and Yang, Jimei and Li, Hao},
  journal = {{arXiv}},
  doi     = {10.48550/arxiv.1812.07035},
  eprint  = {1812.07035}
}

@article{Holden2016, 
  year    = {2016}, 
  title   = {A deep learning framework for character motion synthesis and editing}, 
  author  = {Holden, Daniel and Saito, Jun and Komura, Taku}, 
  journal = {{ACM} Transactions on Graphics ({TOG})}, 
  issn    = {0730-0301}, 
  doi     = {10.1145/2897824.2925975}, 
  pages   = {1--11}, 
  number  = {4}, 
  volume  = {35}
}

@article{Holden2017, 
  year    = {2017}, 
  title   = {Phase-functioned neural networks for character control}, 
  author  = {Holden, Daniel and Komura, Taku and Saito, Jun}, 
  journal = {{ACM} Transactions on Graphics ({TOG})}, 
  issn    = {0730-0301}, 
  doi     = {10.1145/3072959.3073663}, 
  pages   = {1--13}, 
  number  = {4}, 
  volume  = {36}
}

@article{Loshchilov2017, 
  year    = {2017}, 
  title   = {Decoupled Weight Decay Regularization}, 
  author  = {Loshchilov, Ilya and Hutter, Frank}, 
  journal = {{arXiv}}, 
  doi     = {10.48550/arxiv.1711.05101}, 
  eprint  = {1711.05101}
}

@article{Blei2016, 
  year    = {2016}, 
  title   = {Variational Inference: A Review for Statisticians}, 
  author  = {Blei, David M and Kucukelbir, Alp and {McAuliffe}, Jon D}, 
  journal = {{arXiv}}, 
  doi     = {10.48550/arxiv.1601.00670}, 
  eprint  = {1601.00670}
}

@article{Kingma2013, 
  year     = {2013}, 
  title    = {Auto-Encoding Variational Bayes}, 
  author   = {Kingma, Diederik P and Welling, Max}, 
  journal  = {{arXiv} [stat.{ML}]}, 
  issn     = {1312.6114}, 
  doi      = {10.48550/arxiv.1312.6114}, 
  url      = {http://dx.doi.org/10.48550/arxiv.1312.6114}, 
  month    = {12}
}

@article{Rezende2014, 
  year    = {2014}, 
  title   = {Stochastic Backpropagation and Approximate Inference in Deep Generative Models}, 
  author  = {Rezende, Danilo Jimenez and Mohamed, Shakir and Wierstra, Daan}, 
  journal = {{arXiv}}, 
  doi     = {10.48550/arxiv.1401.4082}, 
  eprint  = {1401.4082}
}

@article{Ghosh2023, 
  year     = {2023}, 
  title    = {{ReMoS}: 3D motion-conditioned reaction synthesis for two-person       interactions}, 
  author   = {Ghosh, Anindita and Dabral, Rishabh and Golyanik, Vladislav and Theobalt, Christian and Slusallek, Philipp}, 
  journal  = {{arXiv} [cs.{CV}]}, 
  issn     = {2311.1705}, 
  url      = {http://arxiv.org/abs/2311.17057}, 
  month    = {11}
}

@article{Ji2025, 
  year    = {2025}, 
  title   = {Towards Immersive Human-X Interaction: A Real-Time Framework for Physically Plausible Motion Synthesis}, 
  author  = {Ji, Kaiyang and Shi, Ye and Jin, Zichen and Chen, Kangyi and Xu, Lan and Ma, Yuexin and Yu, Jingyi and Wang, Jingya}, 
  journal = {{arXiv}}, 
  doi     = {10.48550/arxiv.2508.02106}, 
  eprint  = {2508.02106}
}

@article{Jiang2025, 
  year    = {2025}, 
  title   = {{ARFlow}: Human Action-Reaction Flow Matching with Physical Guidance}, 
  author  = {Jiang, Wentao and Wang, Jingya and Ji, Kaiyang and Jia, Baoxiong and Huang, Siyuan and Shi, Ye}, 
  journal = {{arXiv}}, 
  doi     = {10.48550/arxiv.2503.16973}, 
  eprint  = {2503.16973}
}

@article{Garnelo2018z2k, 
  year    = {2018}, 
  title   = {Conditional Neural Processes}, 
  author  = {Garnelo, Marta and Rosenbaum, Dan and Maddison, Chris J and Ramalho, Tiago and Saxton, David and Shanahan, Murray and Teh, Yee Whye and Rezende, Danilo J and Eslami, S M Ali}, 
  journal = {{arXiv}}, 
  doi     = {10.48550/arxiv.1807.01613}, 
  eprint  = {1807.01613}
}

@article{Zhang20244ao, 
  year    = {2024}, 
  title   = {{RoHM}: Robust Human Motion Reconstruction via Diffusion}, 
  author  = {Zhang, Siwei and Bhatnagar, Bharat Lal and Xu, Yuanlu and Winkler, Alexander and Kadlecek, Petr and Tang, Siyu and Bogo, Federica}, 
  journal = {{arXiv}}, 
  doi     = {10.48550/arxiv.2401.08570}, 
  eprint  = {2401.08570}
}

@ARTICLE{Tanke2025-wq,
  title         = "Dyadic Mamba: Long-term dyadic human motion synthesis",
  author        = "Tanke, Julian and Shibuya, Takashi and Uchida, Kengo and
                   Saito, Koichi and Mitsufuji, Yuki",
  journal       = "arXiv [cs.CV]",
  month         =  "14~" # may,
  year          =  2025,
  url           = "http://arxiv.org/abs/2505.09827",
  archivePrefix = "arXiv",
  primaryClass  = "cs.CV",
  doi           = "10.48550/arXiv.2505.09827"
}

@article{Wallot2018, 
  year     = {2018}, 
  title    = {Analyzing Multivariate Dynamics Using Cross-Recurrence Quantification       Analysis ({CRQA}), Diagonal-Cross-Recurrence Profiles ({DCRP}), and
      Multidimensional Recurrence Quantification Analysis ({MdRQA}) - A Tutorial
      in R}, 
  author   = {Wallot, Sebastian and Leonardi, Giuseppe}, 
  journal  = {Frontiers in psychology}, 
  issn     = {1664-1078}, 
  doi      = {10.3389/fpsyg.2018.02232}, 
  url      = {https://www.frontiersin.org/journals/psychology/articles/10.3389/fpsyg.2018.02232/full}, 
  pages    = {2232}, 
  volume   = {9}, 
  month    = {12}
}

@article{Coco2021, 
  year     = {2021}, 
  title    = {Unidimensional and Multidimensional Methods for Recurrence Quantification Analysis with crqa}, 
  author   = {Coco, Moreno,I and Mønster, Dan and Leonardi, Giuseppe and Dale, Rick and Wallot, Sebastian}, 
  journal  = {The R Journal}, 
  issn     = {2006.0195}, 
  doi      = {10.32614/rj-2021-062}, 
  eprint   = {2006.01954}, 
  url      = {https://journal.r-project.org/archive/2021/RJ-2021-062/index.html}, 
  pages    = {145}, 
  number   = {1}, 
  volume   = {13}, 
  month    = {5}
}

@article{Wallot2019, 
  year     = {2019}, 
  title    = {Multidimensional Cross-Recurrence Quantification Analysis ({MdCRQA}) - A       method for quantifying correlation between multivariate time-series}, 
  author   = {Wallot, Sebastian}, 
  journal  = {Multivariate behavioral research}, 
  issn     = {0027-3171}, 
  doi      = {10.1080/00273171.2018.1512846}, 
  url      = {https://www.ncbi.nlm.nih.gov/pubmed/30569740}, 
  pages    = {173--191}, 
  number   = {2}, 
  volume   = {54}, 
  month    = {3}
}

\clearpage

\appendix
\crefalias{section}{appendix}

\section{Derivation of the Evidence Lower Bound}\label{appx:elbo-derivation}

We derive the objective in \cref{eqn:soc-elbo} by introducing the approximate posterior into the
marginal likelihood, applying Jensen's inequality, and separating the temporal and hierarchical
factors. We first take \(\beta_g=\beta_p=1\), then describe the weighted training objective.

\paragraph{Notation and factorisation.}
Let \(Z=(Z^g,Z^{\mathrm{ind}})\) denote the full latent trajectory and
\(H_t=(z_{<t}^g,Z_{<t}^{\mathrm{ind}})\) its history before time~\(t\).
The observed context \(C\) and full target data \(D=(X,Y)\) are fixed throughout the derivation.
To keep the algebra readable, write the conditional densities as
\begin{align*}
  q_t^g &:= q_\phi(z_t^g\mid H_t,C,D),
  & p_t^g &:= p_\theta(z_t^g\mid H_t,C_t), \\
  q_t^p &:= q_\phi(z_t^p\mid z_{<t}^p,z_t^g,C,D),
  & p_t^p &:= p_\theta(z_t^p\mid z_{<t}^p,z_t^g,C_t), \\
  \ell_t^p &:= p_\theta(y_t^p\mid x_t^p,z_t^p).
\end{align*}
As in \cref{eqn:soc-gen-model}, the joint-slot index is suppressed in \(\ell_t^p\); the likelihood
factorises over the target slots for each person. The generative and variational distributions are
therefore
\[
  p_\theta(Y,Z\mid X,C)=\prod_{t=1}^T p_t^g\prod_{p=1}^P p_t^p\ell_t^p,
  \qquad
  q_\phi(Z\mid C,D)=\prod_{t=1}^T q_t^g\prod_{p=1}^P q_t^p.
\]
In particular, there is one group factor per time step and one person factor per person and time step.
Below, \(q\) abbreviates \(q_\phi(Z\mid C,D)\).

\paragraph{From marginal likelihood to a lower bound.}
Multiplying and dividing the integrand by \(q\), and using the concavity of the logarithm, gives
\begin{align}
  \ln p_\theta(Y\mid X,C)
  &= \ln\int p_\theta(Y,Z\mid X,C)\,\mathrm{d}Z \nonumber \\
  &= \ln\mathbb{E}_{q}\!\left[
    \frac{p_\theta(Y,Z\mid X,C)}{q_\phi(Z\mid C,D)}\right] \nonumber \\
  &\geq \mathbb{E}_{q}\!\left[
    \ln\frac{p_\theta(Y,Z\mid X,C)}{q_\phi(Z\mid C,D)}\right]
    =: \mathrm{ELBO}_{1}.
  \label{eqn:elbo-jensen}
\end{align}
The gap is
\(\ln p_\theta(Y\mid X,C)-\mathrm{ELBO}_{1}
=\mathbb{KL}(q_\phi(Z\mid C,D)\|p_\theta(Z\mid X,Y,C))\geq0\),
so equality holds when the variational distribution equals the true posterior.

\paragraph{Separating time, people, and hierarchy.}
Substituting the factorisations into \cref{eqn:elbo-jensen}, expanding the logarithm of each product,
and using linearity of expectation yields
\begin{align}
  \mathrm{ELBO}_{1}
  &= \mathbb{E}_{q}\!\left[
    \ln\frac{\prod_{t=1}^T p_t^g\prod_{p=1}^P p_t^p\ell_t^p}
    {\prod_{t=1}^T q_t^g\prod_{p=1}^P q_t^p}\right] \nonumber \\
  &= \sum_{t=1}^T\sum_{p=1}^P\mathbb{E}_{q}[\ln\ell_t^p]
    -\sum_{t=1}^T\mathbb{E}_{q}\!\left[\ln\frac{q_t^g}{p_t^g}\right]
    -\sum_{t=1}^T\sum_{p=1}^P\mathbb{E}_{q}\!\left[\ln\frac{q_t^p}{p_t^p}\right].
  \label{eqn:elbo-log-ratios}
\end{align}
The first term scores reconstruction of each person's target motion. The remaining terms compare the
target-informed posterior with the context-conditioned prior at the group and person levels.

\paragraph{Conditional KL terms.}
Let \(q_{<t}\) be the marginal distribution of \(H_t\) under \(q\), and let
\(q_{<t,g}\) be the marginal distribution of \((H_t,z_t^g)\).
For a fixed history, integrating the group log-ratio over \(z_t^g\) gives a KL divergence.
For the person log-ratio, we also fix \(z_t^g\) before integrating over \(z_t^p\).
The law of iterated expectation thus gives
\begin{align*}
  \mathbb{E}_{q}\!\left[\ln\frac{q_t^g}{p_t^g}\right]
  &= \mathbb{E}_{q_{<t}}\!\left[
    \mathbb{E}_{q_t^g}\!\left[\ln\frac{q_t^g}{p_t^g}\right]\right]
   = \mathbb{E}_{q_{<t}}[\mathbb{KL}(q_t^g\|p_t^g)], \\
  \mathbb{E}_{q}\!\left[\ln\frac{q_t^p}{p_t^p}\right]
  &= \mathbb{E}_{q_{<t,g}}\!\left[
    \mathbb{E}_{q_t^p}\!\left[\ln\frac{q_t^p}{p_t^p}\right]\right]
   = \mathbb{E}_{q_{<t,g}}[\mathbb{KL}(q_t^p\|p_t^p)].
\end{align*}
Future latents and other current person latents integrate out. The outer expectations remain because
the conditional factors depend on sampled histories and the person factors also depend on the sampled
group latent. Substituting into \cref{eqn:elbo-log-ratios} gives
\begin{align}
  \mathrm{ELBO}_{1}
  &= \sum_{t=1}^T\mathbb{E}_{q}
    [\ln p_\theta(Y_t\mid X_t,z_t^g,Z_t^{\mathrm{ind}})] \nonumber \\
  &\quad-\sum_{t=1}^T\mathbb{E}_{q_{<t}}[\mathbb{KL}(q_t^g\|p_t^g)]
    -\sum_{t=1}^T\sum_{p=1}^P
      \mathbb{E}_{q_{<t,g}}[\mathbb{KL}(q_t^p\|p_t^p)].
  \label{eqn:elbo-decomposed}
\end{align}
Although the likelihood is written using both latent levels, the decoder in
\cref{eqn:soc-gen-model} depends directly on the person latents; the group latent influences it through
those latents. The group KL is counted once per time step, rather than once per person.

\paragraph{Weighting and evaluation during training.}
Denote the expected reconstruction term in \cref{eqn:elbo-decomposed} by \(\mathcal{R}\), and its
summed group and person KL terms by \(\mathcal{K}_g\) and \(\mathcal{K}_p\). Introducing the training
weights gives exactly \cref{eqn:soc-elbo}:
\[
  \mathrm{ELBO}_{\beta}
  =\mathcal{R}-\beta_g\mathcal{K}_g-\beta_p\mathcal{K}_p
  =\mathrm{ELBO}_{1}-(\beta_g-1)\mathcal{K}_g-(\beta_p-1)\mathcal{K}_p.
\]
Thus \(\beta_g=\beta_p=1\) recovers the standard ELBO. If both weights are at least one, the objective
remains a lower bound, but arbitrary weights (for example, values below one during KL warm-up) do not
in general preserve that guarantee. The weights are training choices, not a consequence of Jensen's
inequality. In \cref{eqn:soc-elbo}, the common expectation notation \(\langle\cdot\rangle_{q_\phi}\)
implicitly uses the appropriate marginals specified above.

For the diagonal Gaussian factors used here, each conditional KL can be evaluated analytically at a
sampled history (and sampled group latent for the person terms). The remaining expectations over
latent trajectories and reconstruction can be estimated with reparameterised posterior samples.
The auxiliary reconstruction losses in \cref{eqn:train-loss} are additional supervision and are not
part of this likelihood-bound derivation.

\section{Cross-group person re-identification}\label{appx:z-ind-analysis}

% Source: rebuttal supplement, nps-rbut-13A2/README.md, identity analysis.
\paragraph{Protocol.}
We analyse the model trained jointly on the two-, three-, and four-person subsets of Embody3D.
The analysis split contains \(87\) groups, \(1{,}600\) sequences of \(200\) frames, and \(91\)
individuals. We first average each participant's inferred person latent over the frames of a sequence,
then average across sequences containing that same person in the same group. This produces \(268\)
vectors, one per (person, group) pair. Distances are Euclidean distances in the full latent space;
the two-dimensional PCA projection is used only for visualisation. Identity labels are used only for
evaluation: training includes neither identity supervision nor a re-identification objective.

\paragraph{Cross-group similarity.}
The median same-person distance is \(6.36\), compared with \(16.22\) for different people
(ratio \(0.392\)). To make these distances easier to interpret, we rank each same-person pair within
the different-person distance distribution. The median pair lies at the \(15.7\)th percentile.
Different-person pairs may come from the same group, so the reference distribution also includes
people who share an interaction context.

\begin{figure}[htbp]
  \centering
  \includegraphics[width=1\linewidth]{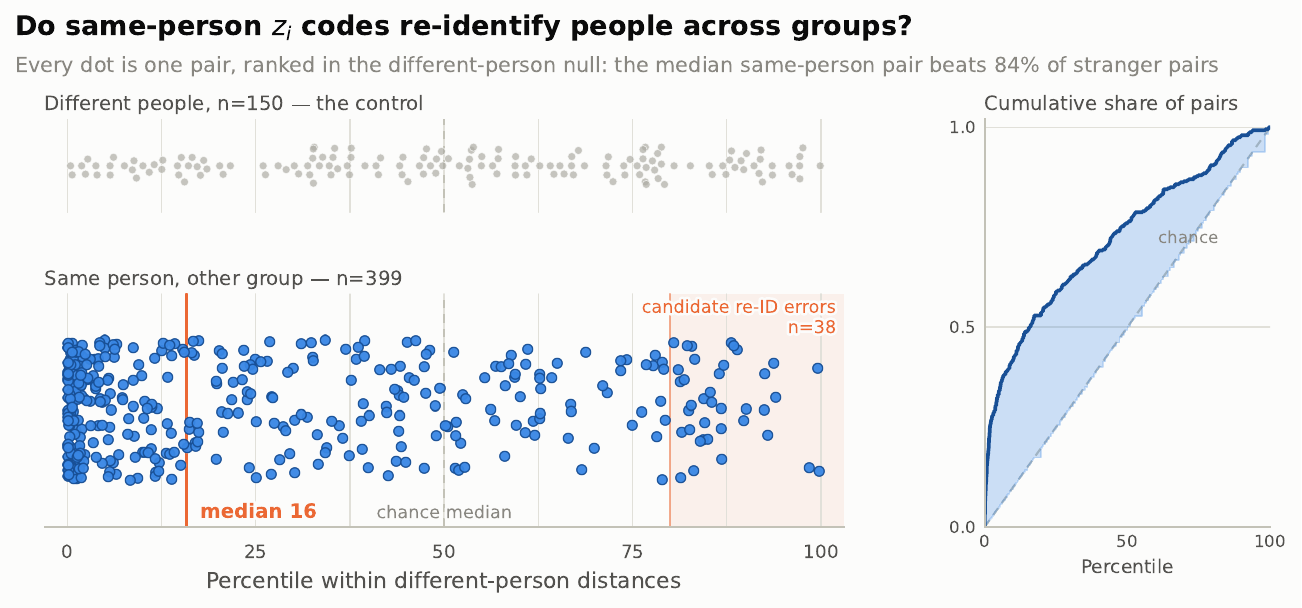}
  \caption{The individual latent \(z^i\) remembers who someone is, across different groups. Each person gets
    one \(z^i\) vector per group they appear in. We measure the distance between two vectors of the same person
    seen in two different groups, and report it as a percentile of the distances between different people:
    percentile \(20\) means that pair is closer together than \(80\%\) of pairs of strangers. If \(z^i\)
    carried no identity-related distance structure, the reference would be a uniform percentile
    distribution with median \(50\).}
  \label{fig:reid-ranking}
\end{figure}

\begin{itemize}
  \item Top left: the displayed different-person control pairs (\(n=150\)) approximately follow the
    uniform percentile reference.

  \item Bottom left: same-person pairs (\(n=399\)) instead pile up near \(0\), with a median percentile of 16:
    the typical person is closer to themselves in another group than \(84\%\) of stranger pairs are to each
    other. The shaded strip holds \(38\) relatively distant same-person pairs above the
    \(80^\text{th}\) percentile, showing that this structure is not equally strong for every pair.

  \item Right: the cumulative view of the same percentiles; the diagonal is the uniform reference,
    and the shaded area shows the excess of close same-person pairs.
\end{itemize}

\paragraph{Nearest-neighbour retrieval.}
For each vector whose person appears in at least two groups, we rank vectors from other groups by distance
and test whether the nearest neighbour belongs to the same individual. This recovers the correct identity
in \(99\) of \(239\) eligible queries (\(41.4\%\) top-1 accuracy), without metric learning or re-ranking.
The \(239\) retrieval queries and \(399\) same-person pairs in \Cref{fig:reid-ranking} count different
units: one is a query vector, the other a cross-group pair. Excluding the query's group tests transfer
across group contexts, although different groups can still share partners or activities. Together with
the distance analysis, retrieval supports persistent participant-specific information in the person
latent, while leaving room for context-dependent variation; it does not establish a context-free or
causally identifiable identity representation.

\section{Training Details}\label{appx:training-details}

\paragraph{Data preparation.}
We train on the Haggling, DnD, DD100, and DuoBox datasets using the
train/validation/test partitions described in the respective papers. For Embody3D we use the split provided
by MAGnet authors. Further, broken clips from Haggling are filtered out.
All datasets are converted to the common motion representation of
\Cref{ssec:motion-repr}. Motion is resampled to 30 fps and divided
into windows of \(200\) frames, using a
stride of \(150\) frames during training. We normalise
all inputs using statistics
computed on the training split only. We apply person order augmentaiton and joint masking.

\paragraph{Training-example construction.}
Each training example contains between \(2\) and \(5\)
participants. To train a single model for forecasting, in-filling, missing-joint
completion, and response generation, we sample a context pattern for every clip.
% Specifically, \emph{[give the probabilities or sampling rule for the task/context patterns]}.
For forecasting examples, the observed prefix contains
\(30\) to \(70\) frames; for temporal in-filling, observations are
retained \emph{[give the interval or mask distribution]}; and for joint- or
person-level completion, we hide \(30\%\) of the joints randomly or completely hide one or more
participants randomly. The observed context varies throughout the clip and across training samples, so the
model learns to adapt to arbitrary context sets. Masks are regenerated at every update with a step-dependent
seed. The target is always the full window of \(200\) frames.

\paragraph{Architecture.}
Unless otherwise stated, models contain approximately \(16\)M trainable parameters. The
group and person latent variables have dimensions \(32\)
and \(16\), respectively, and the person-history LSTM has hidden width
\(256\). The context encoder uses
\(3\) layers with hidden width \(256\), \(8\) attention
heads, and a perceiver resampler. Dedicated heads predict canonical root motion and partner transforms,
while body pose is decoded through a frozen, pretrained motion VAE (\Cref{appx:model-implementation}).
Dropout is set to \(0.1\). Unless noted otherwise, these settings are shared across
all datasets and evaluation tasks.

\paragraph{Reconstruction losses.}
In addition to the ELBO, we supervise the decoded body directly to stabilise training and keep rotations,
joint positions, and global placement mutually consistent. The reconstruction terms in
\cref{eqn:train-loss} use the geodesic distance between rotations on \(\mathrm{SO}(3)\) and the smooth L1
(Huber) penalty:
\begin{equation}
  d_R(\hat{R}, R)=\arccos\bigl((\operatorname{tr}(\hat{R}^{\top} R)-1)/2\bigr),
  \qquad
  \rho(x;\beta)=
  \begin{cases}
    x^2/2\beta, & |x| < \beta, \\
    |x|-\beta/2, & \text{otherwise.}
  \end{cases}
  \label{eqn:geodesic-huber}
\end{equation}
With these, we supervise per-joint rotations (weighted by \(\alpha_j\)), root-relative joint positions, and
the canonical transforms of \(\mathcal{M}\) defined in \cref{ssec:motion-repr}:
{\small
  \begin{align}
    \mathcal{L}_{\mathrm{pose}} &= \mathbb{E}\Bigl[\textstyle\sum_j \alpha_j\,
    \rho\bigl(d_R(\hat{R}^j, R^j);\,1\bigr)\Bigr],
    &\mathcal{L}_{\mathrm{key}} &= \mathbb{E}\bigl[\rho(\hat{K}-K;\,0.05)\bigr], \nonumber \\
    \mathcal{L}_{\mathrm{root}} &= \sum_{m \in \mathcal{M}} \mathbb{E}\bigl[
    \rho\bigl(d_R(\hat{R}_m, R_m);\,1\bigr) + \rho(\hat{t}_m - t_m;\,1)\bigr],
    \label{eqn:recon-terms}
\end{align}}%
Each term is averaged over valid (non-masked) entries and summed over time and people.

\paragraph{Optimisation and objective.}
We optimise all models with AdamW~\citep{Loshchilov2017}, an initial learning rate
of \(1\mathrm{e}-4\) with a cosine scheduler. Weight decay
\(1\mathrm{e}-4\). The base model is trained for
approximately \(50\) passes over the training windows, with the cosine schedule spanning
the full run; gradients are clipped to a maximum norm of \(1\). In
\Cref{eqn:train-loss}, we use
\(\lambda_{\mathrm{pose}}=20\),
\(\lambda_{\mathrm{key}}=10\), and
\(\lambda_{\mathrm{root}}=1\), with unit weight on the rotation and translation of each transform in
\(\mathcal{M}\). We additionally use auxiliary terms with the following weights: joint velocity \(10\)
(Huber, \(\beta=0.05\)); root position \(1\); canonical trajectory position and yaw \(10\) each; canonical
translation and rotation velocity \(10\) and \(5\); rotation-velocity matching \(5\); partner consistency
\(5\); foot contact \(1\); canonical-trajectory negative log-likelihood \(1\); and body negative
log-likelihood \(0.001\).
The group and person KL terms use $\beta_g=0.01$ and
  $\beta_p=0.05$, respectively. During initial training, both
  coefficients are multiplied by $\min(s/1000,1)$, where $s$
  is the optimisation step; continuation runs retain the full
  coefficients without restarting warm-up. The main posterior
  objective applies free bits of $0.05$ nats per latent dimension,
  using $\max(\mathrm{KL}_d,0.05)$ before aggregation. The
  posterior-dropout auxiliary objective uses the same KL
  coefficients without free bits.

\section{Model implementation}\label{appx:model-implementation}

% Source: DD100_TRAINING_STAGES.md (implementation reference, read from saved run configs and code).
Training has two parts: a temporal motion VAE, pretrained once and then frozen, and the hierarchical
latent-variable model, trained on top of the frozen VAE decoder. The hierarchical model is first trained
from scratch and then refined through a sequence of continuation stages (\Cref{tab:training-stages}).

\paragraph{Scope.}
The motion VAE and the staged training below are used for the DD100 and DuoBox models. The Embody3D and
DnD models behind the \(N=2\)--\(5\) results in \Cref{tab:social-forecasting-results} were trained with
the base \methodname{} objective only (\Cref{appx:training-details}), without the pretrained motion VAE or
the continuation stages due to the lack of resources. We will update these results with models trained through the full pipeline.
\textbf{Likewise, the analyses and ablation tables in
\Cref{appx:model-size,appx:additional-baselines,appx:cross-domain,sec:hierarchy-ablation} use
\methodname{} without the motion VAE.}

\subsection{Temporal motion VAE}\label{appx:motion-vae}

\paragraph{Architecture.}
The VAE is a temporal convolutional encoder--decoder of width \(256\) and depth \(3\), with \(4\times\)
temporal compression: each person receives one code \(c_k^p\in\mathbb{R}^{128}\) per block of four
frames. Its input per frame is the \(6\)D rotation of each of the \(J=21\) non-root body joints
together with the canonical-to-root transform \(\mathbf{T}^{\mathrm{can}\to\mathrm{root}}\in\mathbb{R}^9\),
giving \(135\) channels. Encoder and decoder are both conditioned on
\(u_t^p=[\Delta\mathbf{T}_t^{\mathrm{can}},\beta^p]\in\mathbb{R}^{25}\), the canonical frame-to-frame
increment and the \(16\) shape coefficients; the increment of the first frame is set to the identity. Poses, conditions and codes
are standardised with fixed training-set statistics. The encoder outputs a diagonal Gaussian posterior
whose standard deviation is bounded as \(\sigma=0.1+0.9\,\mathrm{sigmoid}(\cdot)\), used consistently for
sampling and for the KL term. A separate head predicts foot-contact logits. During pretraining, the
encoder and decoder see the complete motion of every present person and the true canonical trajectory;
only padding is masked.

\paragraph{Objective.}
With \(\hat{\cdot}\) denoting reconstructions, the VAE minimises
\begin{align}
  \mathcal{L}_{\mathrm{VAE}}
  &= \mathrm{MSE}(\hat\Theta,\Theta)
  + 10\,\lVert \hat K - K\rVert_1
  + 10\,\lVert \hat{t}^{\mathrm{root}} - t^{\mathrm{root}}\rVert_1
  + 0.1\,\rho\bigl(\hat{\dot K}^{\mathrm{w}} - \dot K^{\mathrm{w}};\,0.2\bigr) \nonumber \\
  &\quad + 0.1\,\mathrm{BCE}(\hat{m}, m)
  + 0.5\,\mathcal{L}_{\mathrm{foot}}
  + \beta_{\mathrm{VAE}}\,\mathbb{KL}\bigl(q_\psi(c\mid \Theta,u)\,\|\,\mathcal{N}(0,I)\bigr),
  \label{eqn:vae-loss}
\end{align}
where the MSE is taken over standardised poses, \(K\) are root-relative joint positions and
\(t^{\mathrm{root}}\) root positions (both in metres), \(\dot K^{\mathrm{w}}\) are world-frame joint
velocities, \(\rho\) is the Huber penalty of \cref{eqn:geodesic-huber}, \(m\) are foot-contact labels,
and \(\mathcal{L}_{\mathrm{foot}}\) is a Huber stance/skating penalty. The KL is averaged per latent
dimension, and \(\beta_{\mathrm{VAE}}\) is ramped linearly from \(0\) to \(10^{-3}\) over the first
\(1{,}000\) updates.

\paragraph{Optimisation and freezing.}
We use AdamW with weight decay \(10^{-4}\) and global-norm clipping at \(1\), batch size \(16\), and a
learning rate warmed up from \(10^{-5}\) to \(2\mathrm{e}{-4}\) over \(200\) steps and then cosine-decayed
to \(2\mathrm{e}{-5}\), for approximately \(100\) passes over the training windows. After pretraining, the
VAE weights are excluded from all optimiser updates; gradients still propagate through the frozen
decoder to its inputs. A digest of the VAE parameters is re-verified at the end of every later stage.

\subsection{Coupling the hierarchy to the motion VAE}\label{appx:body-path}

\paragraph{Latent update rates and posterior merging.}
The group latent \(z^g\in\mathbb{R}^{32}\) is updated every five frames and held constant in between;
person latents \(z_t^p\in\mathbb{R}^{16}\) are sampled every frame, conditioned on the current group
sample. Each person carries a recurrent history \(h_t^p\) (LSTM, width \(256\)), and the group transition receives a masked max-pool over people of
features computed from the updated \(h_t^p\) and the previous \(z_{t-1}^p\). In the precision merge of
\cref{eqn:precision-merge}, the prior precision is capped at the bottom-up precision,
\(\sigma_{\mathrm{td}}^{-2}\leftarrow\min(\sigma_{\mathrm{td}}^{-2},\sigma_{\mathrm{bu}}^{-2})\), so the
posterior cannot be dominated by the prior when target evidence is available. During training, the
bottom-up evidence is produced by the context encoder of \cref{ssec:context-enc} applied to the full
target, followed by a backward LSTM smoother.

\paragraph{Trajectory and interaction heads.}
Dedicated heads predict the canonical increments \(\Delta\mathbf{T}_t^{\mathrm{can}}\) and the partner
transforms \(\mathbf{T}_t^{p\to q}\) from \(z^g\), \(z_t^p\), and \(h_t^p\).

\paragraph{Body path.}
Person latents are first smoothed with a fixed \([1,2,1]/4\) temporal filter at strength \(0.5\). A
trainable temporal adapter (four-frame packing, width \(256\), three dilated residual blocks) maps
\([z^g, z^p, \phi(h^p), \bar u^p]\), where \(\phi(h^p)\) are history features and \(\bar u^p\) the
normalised canonical increments and shape coefficients, to standardised \(128\)-dimensional codes
\(\hat c_k^p\). These codes are de-standardised and decoded by the frozen VAE decoder into joint
rotations and \(\mathbf{T}^{\mathrm{can}\to\mathrm{root}}\), which are then supervised with the
reconstruction losses of \cref{eqn:recon-terms}. The decoder condition uses the recorded
\(\Delta\mathbf{T}^{\mathrm{can}}\) with probability \(0.5\) during training (teacher forcing) and the
predicted increment otherwise; at generation it always uses the prediction.

\paragraph{Generation.}
Generation reads only the observed context: latents are sampled from the priors, and neither target
motion nor target codes are accessed. World-frame roots are anchored to the observed roots.

\subsection{Training stages}\label{appx:training-stages}

\begin{table}[htbp]
  \centering
  \caption{Training stages. The motion VAE is frozen after Stage~0. Each continuation restores the full
    state of its parent's final checkpoint.}
  \label{tab:training-stages}
  \small
  \setlength{\tabcolsep}{4pt}
  \renewcommand{\arraystretch}{1.12}
  \begin{adjustbox}{max width=\linewidth}
    \begin{tabular}{@{}l l l p{4.7cm}@{}}
      \toprule
      Stage & Initialisation & Trainable & Change relative to parent \\
      \midrule
      0. Motion VAE & Scratch & VAE & -- \\
      1. Base & Scratch; VAE from 0 & All but VAE & -- \\
      2. Posterior dropout & Stage 1 & All but VAE & Dual-rollout posterior dropout;
      trajectory losses reach \(z^p\) directly \\
      3. Code supervision & Stage 2 & All but VAE & Frozen-encoder code loss on the dense branch \\
      % 4a. Control & Stage 3 & All but VAE & None (Stage 3 recipe continued) \\
      4. VAE-informed posterior & Stage 3 & All but VAE, + projection & Frozen-encoder codes as
      additional \(z^p\) posterior evidence \\
      \bottomrule
    \end{tabular}
  \end{adjustbox}
\end{table}

\paragraph{Stage 1: base model.}
The hierarchical model is initialised from scratch around the frozen VAE and trained with the objective
and schedule of \Cref{appx:training-details} at batch size \(4\). In this stage, gradients from the
trajectory losses to \(z^p\) through its direct input to the trajectory head are stopped.

\paragraph{Continuation recipe (Stages 2--4).}
Each continuation restores model weights, AdamW moments, the step counter, and the random-number state
from its parent's final checkpoint. We verify that a save/restore round trip is bitwise exact and that the
restored model reproduces the parent's final held-out evaluation before training resumes. Continuations
replace the parent's schedule with a constant learning rate of \(3\mathrm{e}{-5}\), keep weight decay and
clipping unchanged, fix the KL multiplier at \(1\) (no annealing), and run for half the number of updates
of Stage~1.

\paragraph{Stage 2: dual-rollout posterior dropout and trajectory coupling.}
Each update evaluates two independent recurrent rollouts with the same weights. The \emph{dense} branch
samples every frame from the posterior and uses the full Stage-1 objective, including teacher forcing and
free bits. The \emph{mixed} branch draws a per-frame mask \(b_t\sim\mathrm{Bernoulli}(0.5)\), shared across
people and the minibatch, and samples frame \(t\) from the posterior if \(b_t=1\) and from the prior
otherwise. Only posterior-sampled frames are scored in this branch, using the likelihood and the raw KL
(the group KL only on its update frames); normalisers use the full count of eligible frames, geometry,
velocity and foot terms are omitted, and the body decoder always uses the predicted trajectory. The two
gradients are summed with unit weight, clipped jointly, and applied as a single AdamW step. In both
branches the posterior retains access to the full target: posterior dropout changes where samples are
drawn from, not the evidence available to the posterior. This stage also removes the Stage-1 stop-gradient,
so trajectory losses update \(z^p\) through its direct input to the trajectory head.

\paragraph{Stage 3: frozen-encoder code supervision.}
In the dense branch only, we add
\begin{equation}
  \mathcal{L}_{\mathrm{code}}
  = \frac{1}{|\mathcal{V}|\,D}\sum_{(b,k,p)\in\mathcal{V}}\sum_{d=1}^{D}
  \left(\frac{\hat c^{\,p}_{b,k,d}-\operatorname{sg}\bigl[\mu_{\psi}(Y,u)^{p}_{b,k,d}\bigr]}{s_d}\right)^{2},
  \label{eqn:code-loss}
\end{equation}
where \(\hat c\) are the adapter's codes in the dense posterior rollout, \(\mu_\psi(Y,u)\) is the frozen VAE
encoder's posterior mean for the complete training motion under the true conditions, \(\operatorname{sg}\)
denotes stop-gradient, \(s_d\) is the fixed training-set standard deviation of code channel \(d\),
\(D=128\), and \(\mathcal{V}\) is the set of valid (batch, code position, person) indices. Targets are
packed exactly as in VAE pretraining. The weight \(\lambda_{\mathrm{code}}=3.5\) was fixed before
training on the first four continuation batches, such that the gradient norm of the code term over the
adapter parameters equals half that of the dense-branch objective. A rule based on the whole-model
gradient norm was not used, since that norm is dominated by trajectory and hierarchy gradients by roughly
three orders of magnitude.

\paragraph{Stage 4: VAE-informed posterior.}
Stage~4 continues Stage~3 with the same recipe and \(\lambda_{\mathrm{code}}\). Frozen-encoder mean codes of the posterior-visible target frames are standardised, repeated
over each four-frame block, passed through a new zero-initialised linear layer
(\(128\to80\), \(10{,}320\) parameters), and added to the per-person evidence entering the bottom-up
person posterior \(q_{\mathrm{bu}}(z_t^p\mid h_t^p,\cdot,z^g)\). The group posterior, all priors, and the
generation path are unchanged; the codes are detached and the VAE remains frozen. Before training we
verify that installing the branch leaves the dense rollout bit-identical, that generation is unchanged
when hidden target frames are corrupted, and that existing AdamW moments are preserved while the new
parameters' moments start at zero.
The DD100 and DuoBox results use the final Stage~4 checkpoint.

\paragraph{Implementation.}
All models are implemented in Jax/Flax NNX.

\section{Model Size Comparisons}\label{appx:model-size}

% Source: author response on model capacity, noteId=mjL4d1kISa.
We vary model capacity while holding the training data, learning-rate schedule, and epoch budget fixed.
This comparison evaluates model selection under a common training budget; it does not establish a
scaling limit. Larger models may require different optimisation settings or more data, which this sweep
does not test.

The \(16\)M model attains the lowest SDTW, cross-person SDTW, and FD on both Panoptic and DD100
(\Cref{tab:model-size-panoptic,tab:model-size-dd100}). Larger variants improve determinism error and
produce greater sample spread, but their higher cross-person SDTW and FD show that this additional
variation does not translate into closer interaction structure or distributional agreement under the
tested setup. The \(16\)M choice therefore provides a useful balance for our emphasis on coordinated,
distributionally faithful generation, rather than a claim of superiority on every diagnostic.

The best result in each column is shown in bold, and the second-best distinct
result is underlined. Tied values receive the same formatting.

\pgfplotstableread{
  Model RR DET SDTW XSDTW FD DIV FS MPJPE MPJVE
  4M  0.011 0.33 3.146 0.774 0.86 0.18 0.52 0.16 0.013
  16M 0.011 0.28 3.061 0.511 0.27 0.42 0.59 0.13 0.012
  25M 0.011 0.21 8.596 1.489 0.34 1.40 0.55 0.20 0.015
  32M 0.011 0.21 8.600 1.371 0.34 1.47 0.61 0.20 0.013
}\panopticmodelsizedata

\begin{table}[htbp]
  \centering
  \caption{Model-size comparison on Panoptic.}
  \label{tab:model-size-panoptic}
  \scriptsize
  \setlength{\tabcolsep}{3pt}
  \renewcommand{\arraystretch}{1.08}
  \begin{adjustbox}{max width=\linewidth}
    \pgfplotstabletypeset[
      columns={Model,RR,DET,SDTW,XSDTW,FD,DIV,FS,MPJPE,MPJVE},
      every head row/.style={before row=\toprule,after row=\midrule},
      every last row/.style={after row=\bottomrule},
      columns/Model/.style={
        string type,column type=l,
        column name={\textbf{Model}}
      },
      columns/RR/.style={
        string type,column type={S[table-format=1.3]},best={min}{\panopticmodelsizedata},
        column name={\multicolumn{1}{c}{\textbf{CRQA (RR)}}}
      },
      columns/DET/.style={
        string type,column type={S[table-format=1.2]},best={min}{\panopticmodelsizedata},
        column name={\multicolumn{1}{c}{\textbf{CRQA (DET)}}}
      },
      columns/SDTW/.style={
        string type,column type={S[table-format=1.3]},best={min}{\panopticmodelsizedata},
        column name={\multicolumn{1}{c}{\textbf{SDTW (\(\times 10^3\))}}}
      },
      columns/XSDTW/.style={
        string type,column type={S[table-format=1.3]},best={min}{\panopticmodelsizedata},
        column name={\multicolumn{1}{c}{\textbf{xPerson SDTW (\(\times 10^3\))}}}
      },
      columns/FD/.style={
        string type,column type={S[table-format=1.2]},best={min}{\panopticmodelsizedata},
        column name={\multicolumn{1}{c}{\textbf{FD}}}
      },
      columns/DIV/.style={
        string type,column type={S[table-format=1.2]},best={max}{\panopticmodelsizedata},
        column name={\multicolumn{1}{c}{\textbf{DIV}}}
      },
      columns/FS/.style={
        string type,column type={S[table-format=1.2]},best={min}{\panopticmodelsizedata},
        column name={\multicolumn{1}{c}{\textbf{FS}}}
      },
      columns/MPJPE/.style={
        string type,column type={S[table-format=1.2]},best={min}{\panopticmodelsizedata},
        column name={\multicolumn{1}{c}{\textbf{MPJPE}}}
      },
      columns/MPJVE/.style={
        string type,column type={S[table-format=1.3]},best={min}{\panopticmodelsizedata},
        column name={\multicolumn{1}{c}{\textbf{MPJVE}}}
      }
    ]{\panopticmodelsizedata}
  \end{adjustbox}
\end{table}

\pgfplotstableread{
  Model RR DET SDTW XSDTW FD DIV FS MPJPE MPJVE
  4M  0.012 0.45 2.818 1.072 2.51 12.00 0.14 0.83 0.04
  16M 0.010 0.36 2.390 0.785 0.63  5.81 0.13 0.72 0.05
  25M 0.012 0.23 6.358 1.526 0.92 33.39 0.16 1.36 0.05
  32M 0.012 0.22 6.202 1.463 1.04 35.47 0.15 1.34 0.05
}\ddhundredmodelsizedata

\begin{table}[htbp]
  \centering
  \caption{Model-size comparison on DD100.}
  \label{tab:model-size-dd100}
  \scriptsize
  \setlength{\tabcolsep}{3pt}
  \renewcommand{\arraystretch}{1.08}
  \begin{adjustbox}{max width=\linewidth}
    \pgfplotstabletypeset[
      columns={Model,RR,DET,SDTW,XSDTW,FD,DIV,FS,MPJPE,MPJVE},
      every head row/.style={before row=\toprule,after row=\midrule},
      every last row/.style={after row=\bottomrule},
      columns/Model/.style={
        string type,column type=l,
        column name={\textbf{Model}}
      },
      columns/RR/.style={
        string type,column type={S[table-format=1.3]},best={min}{\ddhundredmodelsizedata},
        column name={\multicolumn{1}{c}{\textbf{CRQA (RR)}}}
      },
      columns/DET/.style={
        string type,column type={S[table-format=1.2]},best={min}{\ddhundredmodelsizedata},
        column name={\multicolumn{1}{c}{\textbf{CRQA (DET)}}}
      },
      columns/SDTW/.style={
        string type,column type={S[table-format=1.3]},best={min}{\ddhundredmodelsizedata},
        column name={\multicolumn{1}{c}{\textbf{SDTW (\(\times 10^3\))}}}
      },
      columns/XSDTW/.style={
        string type,column type={S[table-format=1.3]},best={min}{\ddhundredmodelsizedata},
        column name={\multicolumn{1}{c}{\textbf{xPerson SDTW (\(\times 10^3\))}}}
      },
      columns/FD/.style={
        string type,column type={S[table-format=1.2]},best={min}{\ddhundredmodelsizedata},
        column name={\multicolumn{1}{c}{\textbf{FD}}}
      },
      columns/DIV/.style={
        string type,column type={S[table-format=2.2]},best={max}{\ddhundredmodelsizedata},
        column name={\multicolumn{1}{c}{\textbf{DIV}}}
      },
      columns/FS/.style={
        string type,column type={S[table-format=1.2]},best={min}{\ddhundredmodelsizedata},
        column name={\multicolumn{1}{c}{\textbf{FS}}}
      },
      columns/MPJPE/.style={
        string type,column type={S[table-format=1.2]},best={min}{\ddhundredmodelsizedata},
        column name={\multicolumn{1}{c}{\textbf{MPJPE}}}
      },
      columns/MPJVE/.style={
        string type,column type={S[table-format=1.2]},best={min}{\ddhundredmodelsizedata},
        column name={\multicolumn{1}{c}{\textbf{MPJVE}}}
      }
    ]{\ddhundredmodelsizedata}
  \end{adjustbox}
\end{table}

\section{Cross-dataset and cross-domain generalisation}\label{appx:cross-domain}

% Source: author response, "Part 1-W1/Q1: Claims vs. Evidence", 30 July 2026.
% https://openreview.net/forum?id=Tqri1CASUu&noteId=lccszuUfg9
\paragraph{Setup.}
We distinguish generalisation to unseen groups from transfer to an unseen activity domain. Model A is
trained jointly on Panoptic Haggling (three-person conversational interaction) and DD100 (partnered dance),
whereas Model B is trained only on DD100. Both are evaluated on the same held-out Panoptic groups, using
\(40\) frames of conversational motion as context and generating the following \(60\) frames. Evaluation
covers \(200\) batches of \(32\) sequences. Thus, both models encounter unseen groups, while conversational
interaction is an unseen training domain only for Model B. Adaptation uses the supplied context without
test-time parameter updates.

\begin{table}[htbp]
  \centering
  \caption{Cross-domain forecasting on held-out Panoptic Haggling groups, with \(40\) context frames and
  \(60\) generated frames. Bold: best; underlined: second best.}
  \label{tab:cross-domain-panoptic}
  \small
  \setlength{\tabcolsep}{3pt}
  \renewcommand{\arraystretch}{1.15}
  \begin{adjustbox}{max width=\linewidth}
    \begin{tabular}{@{}l*{9}{c}@{}}
      \toprule
      Training data & \(\Delta\)RR\(\downarrow\) & \(\Delta\)DET\(\downarrow\) &
      \shortstack{SDTW\(\downarrow\)\\\(\times10^3\)} &
      \shortstack{Cr. SDTW\(\downarrow\)\\\(\times10^3\)} & FD\(\downarrow\) & DIV\(\uparrow\) &
      FS\(\downarrow\) & MPJPE\(\downarrow\) & MPJVE\(\downarrow\) \\
      \midrule
      Panoptic + DD100 & \textbf{0.011} & \textbf{0.28} & \textbf{3.061} & \textbf{0.511} &
      \textbf{0.27} & \underline{0.42} & \underline{0.59} & \textbf{0.13} & \textbf{0.012} \\
      DD100 only & \textbf{0.011} & \underline{0.31} & \underline{4.473} & \underline{1.424} &
      \underline{1.94} & \textbf{1.10} & \textbf{0.24} & \underline{0.28} & \underline{0.026} \\
      \bottomrule
    \end{tabular}
  \end{adjustbox}
\end{table}

\begin{figure}[htbp]
  \centering
  \input{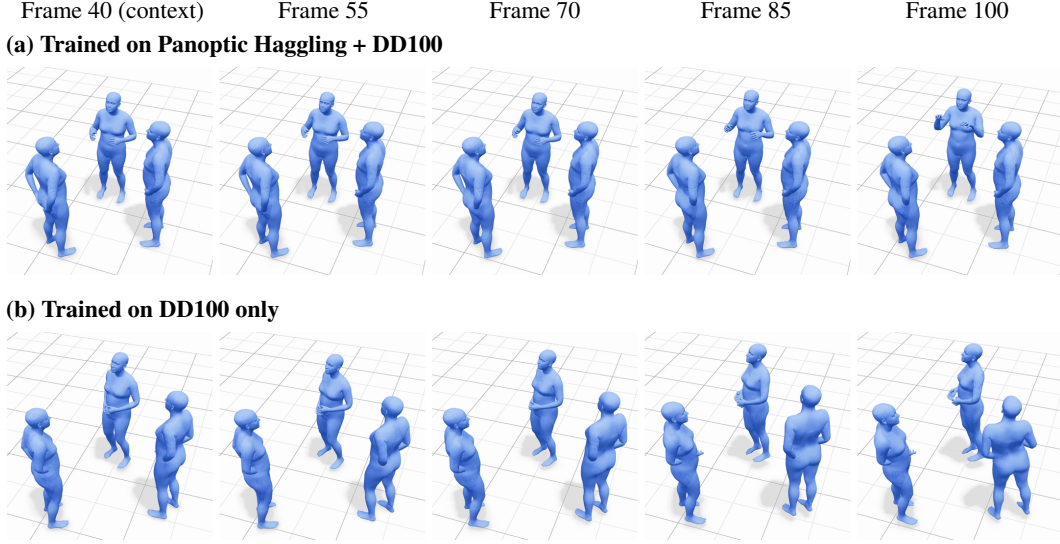}
  \caption{Cross-domain forecasting on a held-out Panoptic Haggling group, using frames from the
    supplementary videos. Both models observe the same \(40\) frames of conversational motion and
    generate the next \(60\). Context frames are rendered from each model's own reconstruction, so the
    frame-\(40\) panels differ slightly between rows. (a) The jointly trained model keeps a stable
    conversational formation while the central participant continues to gesture. (b) The DD100-only model, which never sees
    conversational data during training, also preserves plausible poses and a coherent conversational
    formation, indicating that it carries forward the group structure given in the context. However, the
    right-hand participant turns away from the group, a rotation characteristic of partnered dance rather
    than conversation.}
  \label{fig:cross-domain-examples}
\end{figure}

\paragraph{Activity-specific dynamics within a shared model.}
Joint training improves cross-person SDTW from \(1.424\) to \(0.511\) and FD from \(1.94\) to \(0.27\)
(\Cref{tab:cross-domain-panoptic}), alongside closer temporal alignment and reference recovery. In the
qualitative comparison (\Cref{fig:cross-domain-examples}), the jointly trained model maintains the
conversational formation and generates gestures appropriate to that activity. The dance-only model retains
plausible body poses and a sensible
group formation, but also produces spinning motions characteristic of its training domain. This separates
transfer of broad pose and formation structure from adaptation to activity-specific interaction dynamics.
The result supports learning multiple domains with shared weights and conditioning on new groups; it does
not establish equivalent performance on activities absent from training.

The dance-only model has higher DIV and lower FS, illustrating why sample spread and foot stability should
be interpreted alongside activity-appropriate alignment and distributional fit. The group-latent structure
in \Cref{fig:zg-emb} complements this comparison: corpus and dance-family organisation is consistent with
the inferred social state retaining activity information within a shared model.

\section{Contribution of the bilevel hierarchy}\label{sec:hierarchy-ablation}

% Source: author response, "Part 3-W3/Q3: Ablations for the bilevel hierarchy", 30 July 2026.
% https://openreview.net/forum?id=Tqri1CASUu&noteId=a8seiTZXqk
% The response reports a complete Panoptic table and selected DD100 results in prose.
% It does not explicitly specify the context length or horizon for this ablation.
\paragraph{Variants.}
\textbf{All variants in this section, including the full model, are trained without the motion VAE
(\Cref{appx:model-implementation}).} We compare the full group-and-person hierarchy with two removals, keeping the remainder of the architecture
fixed. The \emph{group-only} variant removes person latents, so individual variation must be represented
through the shared \(z^g\). The \emph{person-only} variant removes social context: the person axis is folded
into the batch, and each participant is modelled independently through \(z^p\). These comparisons test the
functional contributions of the two levels alongside the latent-space analysis in
\Cref{appx:z-ind-analysis}.

\begin{table}[htbp]
  \centering
  \caption{Bilevel-hierarchy ablations on Panoptic Haggling. Bold: best; underlined: second best.}
  \label{tab:hierarchy-ablation-panoptic}
  \small
  \setlength{\tabcolsep}{3pt}
  \renewcommand{\arraystretch}{1.15}
  \begin{adjustbox}{max width=\linewidth}
    \begin{tabular}{@{}l*{9}{c}@{}}
      \toprule
      Latent structure & \(\Delta\)RR\(\downarrow\) & \(\Delta\)DET\(\downarrow\) &
      \shortstack{SDTW\(\downarrow\)\\\(\times10^3\)} &
      \shortstack{Cr. SDTW\(\downarrow\)\\\(\times10^3\)} & FD\(\downarrow\) & DIV\(\uparrow\) &
      FS\(\downarrow\) & MPJPE\(\downarrow\) & MPJVE\(\downarrow\) \\
      \midrule
      \(z_g\) and \(z_i\) & \textbf{0.011} & \underline{0.28} & \textbf{3.061} & \textbf{0.511} &
      \underline{0.27} & 0.42 & \textbf{0.59} & \textbf{0.13} & \textbf{0.012} \\
      \(z_g\) & \textbf{0.011} & 0.38 & \underline{3.193} & \underline{1.119} &
      0.51 & \underline{0.67} & 0.69 & 0.21 & \textbf{0.012} \\
      Person only (no social context) & \underline{0.012} & \textbf{0.19} & 8.530 & 1.305 &
      \textbf{0.18} & \textbf{1.28} & \underline{0.61} & \underline{0.18} & \underline{0.014} \\
      \bottomrule
    \end{tabular}
  \end{adjustbox}
\end{table}

\paragraph{Individual flexibility supports interpersonal alignment.}
On Panoptic, removing person latents changes SDTW modestly, from \(3.061\) to \(3.193\), but more than
doubles cross-person SDTW, from \(0.511\) to \(1.119\)
(\Cref{tab:hierarchy-ablation-panoptic}). Removing social context increases cross-person SDTW further to
\(1.305\), while SDTW rises to \(8.530\). The same cross-person ordering appears on DD100: \(0.785\) for
the full hierarchy, \(0.920\) for group only, and \(1.534\) for person only. On DD100, removing social
context also raises SDTW from \(2.39\) to \(6.35\) and MPJPE from \(0.72\) to \(1.30\). Together, these
results suggest that separate person latents help express individual behaviour in relation to the group,
and that social context contributes to both interpersonal alignment and individual motion prediction.

\paragraph{Complementary metrics reveal the contribution of social structure.}
The person-only model attains lower FD on Panoptic (\(0.18\) versus \(0.27\)) and lower determinism error
on both datasets. These are complementary strengths rather than evidence of uniform superiority by either
variant. FD over per-person frame features evaluates marginal motion statistics, while determinism
summarises one aspect of recurrence structure; neither alone establishes that the generated participants
remain aligned with one another. The full hierarchy's advantage in cross-person SDTW therefore provides
more specific evidence for the social-state motivation: modelling participants jointly helps recover the
structure of their interaction. These architectural ablations support distinct functional contributions
of group and person latents, without asserting causal identifiability of the learned representation.

% Source: anonymous rebuttal supplement nps-rbut-13A2 and the author response describing
% the group-latent source switch after 50 predicted frames.
\paragraph{Conditional person-level variation.}
We also examine generation while keeping the observed context and group latent fixed and varying only
the person latents (\Cref{fig:latent-generation-examples}a). In the DuoBox example, three samples
produce different poses and orientations from the same context. Their relative arrangements also vary:
the person latents express individual realisations of an interaction rather than motion details that
leave group geometry invariant. This complements the re-identification result in \Cref{appx:z-ind-analysis}:
participant-specific information can persist across contexts while allowing multiple ways to participate
in the current interaction.

\paragraph{Changing the group state during generation.}
In a second experiment, we generate \(50\) frames using a group-latent trajectory from one source
sequence, then switch to the trajectory from another sequence containing the same individuals in a
different interaction. Person latents are sampled conditionally on the active group latent, so they
respond to the switch through the hierarchy. The selected frames in \Cref{fig:latent-generation-examples}b
show a change in participants' postures and group arrangement. This illustrates that the group latent
affects the generated interaction through its conditional person-level realisations. Since the person
latents also change, the experiment tests the hierarchy's response to a change of group state, rather
than isolating a direct effect of the group latent with every other latent held fixed.

\begin{figure}[!htb]
  \centering
  \input{figs/latent_generation_examples}
  \caption{Qualitative probes of the latent hierarchy, using frames from the supplementary videos.
    (a) Three generated dyads (blue) share the same context (peach, right) and group latent but use
    different person-latent samples. (b) Selected states before and after changing the group-latent
  source; person latents are sampled conditionally on the active group state.}
  \label{fig:latent-generation-examples}
\end{figure}

\section{Evaluation metrics and interpretation}\label{appx:evaluation-details}

% Source: author response on metrics, noteId=miG9qkGu9y.
Our evaluation follows the multiparty behaviour framework of \citet{Shirekar2025} and combines
interaction structure, distributional agreement, sample variation, and physical diagnostics. These
quantities address complementary questions; no single score establishes that a continuation is both
plausible and socially appropriate.

\paragraph{Baselines.}
RS and NN are naive, non-learned reference baselines: RS returns a randomly sampled sequence from the
training data, and NN returns the training sequence closest to the observed context. They indicate the
scores reachable by replaying recorded motion without modelling the interaction, and are therefore excluded
from the best/second-best ranking in the main tables. MAGNet~\citep{Maluleke2025} was not evaluated on
five-person interactions in its original work, so we do not report it for DnD (\(N=5\)) in
\Cref{tab:social-forecasting-results}; for that setting under full observation, where no learned baseline
remains, the ranking includes RS and NN.

\paragraph{Recurrence and temporal coordination.}
Cross-recurrence quantification analysis (CRQA) compares participants' motion states across time.
A recurrence occurs when two states are within a dataset-specific distance radius, calibrated to a
\(2\%\) recurrence rate and then held fixed for analysis~\citep{Wallot2018, Shirekar2025, Coco2021, Wallot2019}. Recurrence rate (RR) is the fraction of
state pairs marked as recurrent. Determinism (DET) is the fraction of recurrent points that belong to
diagonal line structures, reflecting sustained sequences of similar states. Off-diagonal recurrences
allow temporal offsets, so the comparison can capture leading and following rather than requiring
participants to move identically at the same instant. We report absolute deviations of RR and DET
from their ground-truth values, denoted \(\Delta\mathrm{RR}\) and \(\Delta\mathrm{DET}\); lower values
indicate closer agreement with the recorded recurrence statistics.

\paragraph{Trajectory structure with timing flexibility.}
Soft dynamic time warping (SDTW) compares motion trajectories while allowing temporal alignment.
It therefore tolerates timing shifts that framewise position error penalises directly. The cross-person
variant applies this comparison to relations between participants' motion signals, testing whether the
generated interaction preserves their relative temporal structure. Lower values are better in both
cases. Cross-person SDTW is especially relevant to our social-state motivation: individually plausible
motions can still form a poorly aligned group. Where a table header specifies \(\times10^3\), each displayed entry represents
that many thousands in the underlying SDTW score.

\paragraph{Distributional fit and variation.}
Fr\'echet distance (FD) compares generated and reference motion distributions through the means and
covariances of their evaluation features. Lower FD indicates closer agreement in that feature space.
For the hierarchy ablations, these are per-person frame features: a good marginal motion distribution
can coexist with errors in how people move together. Diversity (DIV) measures the average pairwise
distance between generated samples. Higher DIV indicates more variation, but does not by itself
demonstrate coverage of valid interactions. We consequently interpret DIV alongside FD, temporal
structure, and the activity: a relatively stationary conversation need not exhibit the same motion
spread as dancing or boxing.

\paragraph{Reference recovery and physical artefacts.}
Mean per-joint position error (MPJPE) averages the distance between predicted and recorded joint
positions; mean per-joint velocity error (MPJVE) similarly measures velocity disagreement. Lower values
indicate closer recovery of the particular recorded continuation. Because other continuations may also
be valid, these errors complement the generative metrics rather than defining overall sample quality.
Foot skating (FS) measures sliding during detected ground contact, and interpenetration (IP) measures
body overlap. Lower values indicate fewer of these specific artefacts; they provide useful physical
checks without establishing physical plausibility in every respect. Both are also non-zero in recorded motion
for some activities, for example when dancers in close hold overlap under the capsule test or boxers move
their feet quickly, so values below the recorded rate need not indicate better motion. Where relevant, we
quote these recorded rates alongside the results.

\section{Additional neural-process baseline comparisons}\label{appx:additional-baselines}

% Source: author response, "Part 2-W2: Comparisons do not always favour BRAID", 31 July 2026.
% https://openreview.net/forum?id=Tqri1CASUu&noteId=EHf9QJEtKt
\paragraph{Protocol and baselines.}
We report an additional forecasting comparison using \(70\) observed frames with \(20\%\) of joints
randomly missing and a \(130\)-frame prediction horizon. These results belong to this specific protocol
and should be read separately from the main evaluation tables. We compare \methodname{} with
SP-style~\citep{Raman2023xgd} and SNP-style~\citep{Singh2019-qw} implementations and
RoHM~\citep{Zhang20244ao}. The SP-style implementation is adapted to accept missing joints;
the SNP-style variant uses a single group latent without the person-level hierarchy.

\begin{table}[tbp]
  \centering
  \caption{Additional baselines on Panoptic Haggling: \(70\) observed and \(130\) predicted frames,
  with \(20\%\) missing context joints. Bold: best; underlined: second best; --: unreported.}
  \label{tab:additional-baselines-panoptic}
  \small
  \setlength{\tabcolsep}{3pt}
  \renewcommand{\arraystretch}{1.15}
  \begin{adjustbox}{max width=\linewidth}
    \begin{tabular}{@{}l*{9}{c}@{}}
      \toprule
      Model & \(\Delta\)RR\(\downarrow\) & \(\Delta\)DET\(\downarrow\) &
      \shortstack{SDTW\(\downarrow\)\\\(\times10^3\)} &
      \shortstack{Cr. SDTW\(\downarrow\)\\\(\times10^3\)} & FD\(\downarrow\) & DIV\(\uparrow\) &
      FS\(\downarrow\) & MPJPE\(\downarrow\) & MPJVE\(\downarrow\) \\
      \midrule
      \methodname{} & \textbf{0.011} & 0.28 & \textbf{3.061} & \textbf{0.511} & \textbf{0.27} &
      \underline{0.42} & 0.59 & \textbf{0.13} & \underline{0.012} \\
      SNP-style & \textbf{0.011} & 0.38 & \underline{3.193} & 1.119 & \underline{0.51} & \textbf{0.67} & 0.69
      & \underline{0.21} & \underline{0.012} \\
      SP-style & \textbf{0.011} & \textbf{0.20} & 4.530 & 1.940 & 0.92 & 0.11 & \underline{0.31} &
      \underline{0.21} & \textbf{0.006} \\
      RoHM & \underline{0.013} & \underline{0.21} & 8.251 & \underline{1.007} & 1.285 & -- & \textbf{0.17} &
      0.442 & 0.052 \\
      \bottomrule
    \end{tabular}
  \end{adjustbox}

  \vspace{0.8em}
  \caption{Additional baselines on DD100: \(70\) observed and \(130\) predicted frames,
  with \(20\%\) missing context joints. Bold: best; underlined: second best; --: unreported.}
  \label{tab:additional-baselines-dd100}
  \begin{adjustbox}{max width=\linewidth}
    \begin{tabular}{@{}l*{9}{c}@{}}
      \toprule
      Model & \(\Delta\)RR\(\downarrow\) & \(\Delta\)DET\(\downarrow\) &
      \shortstack{SDTW\(\downarrow\)\\\(\times10^3\)} &
      \shortstack{Cr. SDTW\(\downarrow\)\\\(\times10^3\)} & FD\(\downarrow\) & DIV\(\uparrow\) &
      FS\(\downarrow\) & MPJPE\(\downarrow\) & MPJVE\(\downarrow\) \\
      \midrule
      \methodname{} & \textbf{0.010} & 0.36 & \textbf{2.390} & \textbf{0.785} & \textbf{0.63} & \textbf{5.81}
      & 0.13 & \underline{0.72} & 0.05 \\
      SNP-style & \underline{0.012} & 0.38 & \underline{2.620} & \underline{0.920} & \underline{1.45} &
      \underline{3.53} & \underline{0.12} & \textbf{0.71} & \underline{0.04} \\
      SP-style & \textbf{0.010} & \textbf{0.28} & 4.303 & 1.504 & 2.07 & 1.28 & \textbf{0.08} & 0.90 & \textbf{0.03} \\
      RoHM & 0.013 & \underline{0.29} & 7.017 & 1.493 & 2.29 & -- & 0.31 & 1.03 & 0.11 \\
      \bottomrule
    \end{tabular}
  \end{adjustbox}
\end{table}

\paragraph{A consistent gain in trajectory and interaction structure.}
\methodname{} achieves the lowest SDTW, cross-person SDTW, and FD on both datasets in this comparison
(\Cref{tab:additional-baselines-panoptic,tab:additional-baselines-dd100}). The result links the hierarchy's
benefit to both individual trajectory structure and interpersonal alignment, alongside better
distributional agreement. The advantage is not uniform across all diagnostics: SP-style has lower
determinism and velocity errors, and lower foot skating; SNP-style yields greater diversity on Panoptic
and slightly lower position error on DD100. On DD100, \methodname{} combines the highest reported
diversity with the lowest FD. Read together, these measures support richer samples that remain close
to the reference distribution under this protocol, while the physical and reconstruction scores
identify complementary strengths of the baselines.

\end{document}